\pdfoutput=1
\documentclass[11pt]{article}

\usepackage[final]{acl}

\usepackage{times}
\usepackage{latexsym}

\usepackage[T1]{fontenc}
\usepackage[utf8]{inputenc}
\usepackage{multirow}
\usepackage{graphicx}
\usepackage{algorithm}
\usepackage{algpseudocode}
\usepackage{amsmath}
\usepackage{pifont}
\usepackage{booktabs}
\usepackage{rotating}
\usepackage{listings}
\usepackage{tcolorbox}
\usepackage{CJKutf8}
\usepackage[T1]{fontenc}

\usepackage{microtype}
\usepackage{hyperref}
\tcbuselibrary{listings} % 如果你需要listings库的功能
\newtcolorbox[list inside=prompt,auto counter]{prompt}[1][]{
    colbacktitle=black!60,
    coltitle=white,
    fontupper=\footnotesize,
    boxsep=5pt,
    left=0pt,
    right=0pt,
    top=0pt,
    bottom=0pt,
    boxrule=1pt,
    #1,
}

\usepackage{microtype}

\usepackage{inconsolata}

\usepackage{graphicx}

\title{TableEval: A Real-World Benchmark for \\ Complex, Multilingual, and Multi-Structured Table Question Answering}

\author{First Author \\
  Affiliation / Address line 1 \\
  Affiliation / Address line 2 \\
  Affiliation / Address line 3 \\
  \texttt{email@domain} \\\And
  Second Author \\
  Affiliation / Address line 1 \\
  Affiliation / Address line 2 \\
  Affiliation / Address line 3 \\
  \texttt{email@domain} \\}

\author{
  Junnan Zhu$^{2}$\thanks{Equal Contribution.},
  Jingyi Wang$^{1,2}$\footnotemark[1],
  Bohan Yu$^{1,3}$,
  Xiaoyu Wu$^{1}$,
  Junbo Li$^{1}$,
  Lei Wang$^{1,2}$, 
  Nan Xu$^{1,2}$\thanks{Corresponding Author.}
  \\
  $^1$ Beijing Wenge Technology Co., Ltd. \\
  $^2$ State Key Laboratory of Multimodal Artificial Intelligence Systems, \\Institute of Automation, CAS, Beijing, China \\
  $^3$ School of Advanced Interdisciplinary Sciences, University of Chinese Academy of Sciences, China \\ 
  {\texttt junnan.zhu@nlpr.ia.ac.cn, nan.xu@wenge.com}\\
}

\begin{document}
\maketitle
\begin{abstract}
LLMs have shown impressive progress in natural language processing. However, they still face significant challenges in TableQA, where real-world complexities such as diverse table structures, multilingual data, and domain-specific reasoning are crucial. Existing TableQA benchmarks are often limited by their focus on simple flat tables and suffer from data leakage. Furthermore, most benchmarks are monolingual and fail to capture the cross-lingual and cross-domain variability in practical applications. To address these limitations, we introduce TableEval, a new benchmark designed to evaluate LLMs on realistic TableQA tasks. Specifically, TableEval includes tables with various structures (such as concise, hierarchical, and nested tables) collected from four domains (including government, finance, academia, and industry reports). Additionally, TableEval features cross-lingual scenarios with tables in Simplified Chinese, Traditional Chinese, and English. To reduce potential data leakage, we curate data from recent real-world documents. Considering that existing TableQA metrics fail to capture semantic accuracy, we further propose SEAT, a new evaluation framework that assesses the alignment between model responses and reference answers at the sub-question level. Experimental results have shown that SEAT achieves high agreement with human judgment. Extensive experiments on TableEval reveal critical gaps in the ability of state-of-the-art LLMs to handle these complex, real-world TableQA tasks, offering insights for future improvements. We make our dataset available here: \url{https://github.com/wenge-research/TableEval}. 
\end{abstract}

\section{Introduction}

In recent years, Large Language Models (LLMs) have made remarkable progress in natural language processing tasks such as text generation, reasoning, and question answering. However, the complexity of table-based Question Answering (TableQA) poses distinctive challenges. Unlike texts, tables often exhibit nested structures, multi-row/column spans, and cross-references that demand domain-specific and structure-aware reasoning. 

Despite significant advances in TableQA, current benchmarks fail to capture the key challenges encountered in industrial and real-world applications. First, most~\cite{Chenwh20TabFact, parikh-etal-2020-totto, chen2021open, nan-etal-2022-fetaqa, wu2024tablebench} focus on simple flat tables, ignoring the diverse and complex structures commonly encountered in real-world data, such as hierarchical, concise, and nested structures. 
Second, existing benchmarks suffer from data leakage, where test data overlaps with the pretraining corpora of LLMs. This issue leads to a bias in performance evaluation, producing misleading results that fail to reflect actual generalization ability.
Finally, the predominantly monolingual nature of most benchmarks fails to address the cross-lingual and cross-domain challenges critical for real-world applications. Practical use cases often require the ability to understand and reason over data presented in multiple languages and originating from diverse domains.

To address these limitations, we introduce TableEval, a new benchmark to evaluate LLMs on broader and more representative tasks. TableEval contains diverse tables covering various structures, such as concise\footnote{We refer to concise tables as defined in \citet{webtable2013}, where merged cells are used to enhance compactness.}, hierarchical, and nested tables. We minimize the risks of data leakage by ensuring that all data is sourced from recent documents (published in 2024) that have not been included in the pretraining data of most existing LLMs. Besides, we highlight the challenges of both cross-lingual and cross-domain settings by including tables from various domains (such as administrative records, financial disclosures, academic papers, and industry reports) in multiple languages (Simplified and Traditional Chinese, English), with all QA pairs presented in Simplified Chinese.

TableEval covers six high-level tasks and 16 fine-grained question types, ranging from simple lookup queries to numerical calculations, comparative analysis, and multi-hop questions. The benchmark also includes multi-turn conversation tasks, where models are required to perform dynamic reasoning across multiple steps. Critically, all questions have unambiguous, objectively verifiable answers to ensure consistent and reliable evaluation. 

Evaluating LLMs in TableQA tasks is challenging, primarily because LLMs often generate natural language responses that vary significantly in phrasing and structure. Even when specific output formats are enforced via prompting, LLMs tend to deviate, making it difficult to evaluate the semantic accuracy of responses using traditional metrics like exact match~\cite{zhu-etal-2021-tat}, F1 score~\cite{chen-etal-2021-finqa}, or n-gram matching~\cite{parikh-etal-2020-totto, nan-etal-2022-fetaqa}. To address this limitation, we propose SEAT (\textbf{S}tructured \textbf{E}valuation for \textbf{A}nswers in \textbf{T}ableQA), a novel TableQA evaluation framework that leverages an LLM with crafted prompts to compare generated responses against structured reference answers. SEAT evaluates the semantic correctness of responses by following a two-step process: (1) Extract key answers to each sub-question from the model's response and compare each with the reference to identify matching or divergent elements. If no multiple sub-questions are provided, the original question is treated as a single sub-question. (2) Present the evaluation results in a structured JSON format and aggregate the final scores, facilitating easy tracking and verification. This approach provides a reliable, scalable evaluation method that goes beyond surface-level matching, offering a more accurate assessment.

We systematically assess LLMs on TableEval. Our extensive experiments reveal several notable observations: (1) Closed-source models (e.g., o1-preview, Claude 3.5 Sonnet) consistently lead in performance, but large open-source models (e.g., DeepSeek-R1, QwQ-32B-Preview) show significant promise with proper scaling and enhanced reasoning capabilities. (2) Understanding complex table structures remains a fundamental challenge, with performance drops of 10--15\% on nested/hierarchical tables compared to flat tables. (3) Domain-specific and language-specific gaps persist, as models often struggle more in Chinese or specialized domains than in English or general settings. 

Our main contributions are as follows:

\begin{itemize} 
	\item We introduce TableEval, a comprehensive real-world benchmark with diverse table structures and cross-lingual/cross-domain challenges using recent data to prevent leakage and realistically evaluate LLMs' generalization capabilities.
	\item We propose SEAT, a novel evaluation framework that assesses model responses at the sub-question level, ensuring semantic alignment with reference answers. Through extensive experiments on multiple LLM backbones, SEAT demonstrates consistently higher correlations with human judgments than existing metrics.
	\item Our experiments reveal critical performance gaps in real-world TableQA: closed-source models lead overall but open-source alternatives become competitive with sufficient scaling and enhanced reasoning; complex table structures significantly degrade performance; and domain/language-specific weaknesses indicate the need for more effective strategies.

\end{itemize}

\begin{figure*}[t]
\setlength{\belowcaptionskip}{-0.5cm}
  \centering 
  \includegraphics[width=1.0 \linewidth]{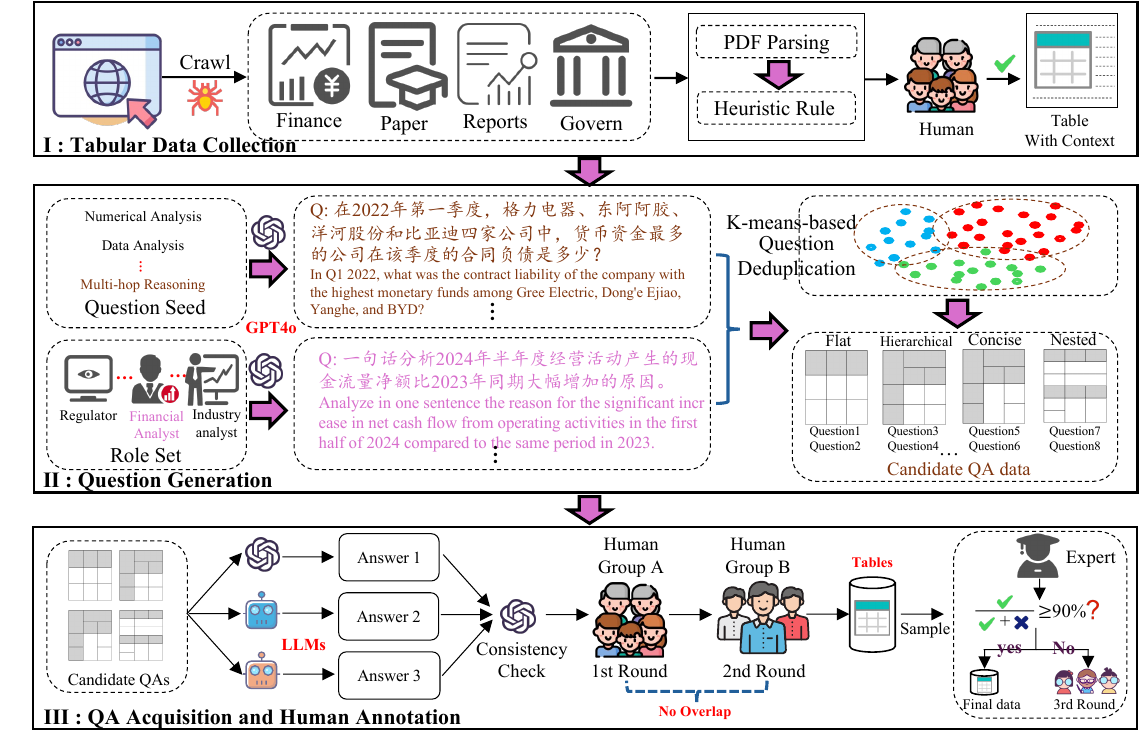}\\
  \caption{Overview of data collection. (1) Tabular Data Collection, collecting tables from financial reports, industry research, academic papers, and governmental data; (2) Question Generation, using Template-Prompted and Role-Prompted strategies to generate TableQA questions, filtered through clustering and deduplication; (3) QA Acquisition and Human Annotation—iteratively refining answers through LLM consistency checks, human reviews, and structured answer extraction, ensuring accuracy, completeness, and alignment with the original tabular data.\label{data}}
\end{figure*}

\section{Related Work}

\textbf{TableQA Methods.} Question answering over tabular data has attracted increasing interest due to its potential to make complex data more accessible to non-experts. Early studies on TableQA primarily explored semantic parsing methods, treating the task as Text-to-SQL, where natural language queries and table schemas are directly converted into executable SQL~\cite{pasupat-liang-2015-compositional, wang-etal-2020-rat, shi-etal-2020-potential, guo-etal-2021-chase}. While these methods show great potential, they require a large amount of SQL-labeled training data and often assume that column headers are enough for reasoning, ignoring the valuable information in table cells. To overcome these problems, recent studies have shifted towards the retrieve-and-reasoning paradigm~\cite{eisenschlos-etal-2021-mate, ijcai2022p0629, kumar-etal-2023-multi, lei-etal-2023-s3hqa} instead of directly generating SQL. Large-scale table pre-training~\cite{yin-etal-2020-tabert, herzig-etal-2020-tapas, liu2022tapex, jiang-etal-2022-omnitab, zhao-etal-2022-reastap} has further enhanced this shift, which has demonstrated strong performance on standard benchmarks. Furthermore, the advent of LLMs enhanced with few-shot prompting~\cite{zhang-etal-2024-tablellama, reactable2024, wang2024chainoftable, sui2024table, li-2024-gpt} has significantly extended these capabilities, enabling more flexible, scalable, and generalizable reasoning over tabular data.

\textbf{TableQA Benchmarks.} While TableQA solutions continue to advance, existing benchmarks have yet to fully catch up with this progress. Early benchmarks like WQA~\cite{pasupat-liang-2015-compositional}, SQA~\cite{iyyer-etal-2017-search}, and TabFact~\cite{Chenwh20TabFact} use HTML tables from Wikipedia, focusing on cell retrieval and extraction tasks with relatively simple table structures.
To address more complex reasoning, later benchmarks introduce greater diversity and difficulty. Datasets such as ToTTo~\cite{parikh-etal-2020-totto}, OTTQA~\cite{chen2021open}, and FeTaQA~\cite{nan-etal-2022-fetaqa} require generating answers beyond the table content, while numeric-focused datasets like FinQA~\cite{chen-etal-2021-finqa} and AIT-QA~\cite{katsis-etal-2022-ait} emphasize computational reasoning in financial contexts. Additionally, logical expression-based benchmarks like Spider~\cite{yu-etal-2018-spider} and Bird~\cite{li2023can} incorporate structured logical reasoning for handling intricate, logic-driven queries.
However, existing benchmarks have the following limitations. Most focus on simple flat table structures, ignoring the hierarchical and nested structures encountered in the real world. Another challenge is data leakage, where overlaps between test data and pretraining corpora lead to biased evaluation. Additionally, most benchmarks are monolingual, lacking the cross-lingual and cross-domain diversity required for real-world applicability.

Therefore, we introduce a comprehensive benchmark that covers various table structures, supports multiple languages, and spans diverse real-world tasks and question types. It offers a more accurate evaluation of TableQA systems. We compare TableEval with existing benchmarks in \autoref{table-comparison}.

\begin{table*}[t]
\small
\centering
\setlength{\belowcaptionskip}{-0.4cm}
\resizebox{\linewidth}{!}{
\begin{tabular}{@{}lccccccccccccc@{}}
\toprule
Dataset    & Flat      & \begin{tabular}[c]{@{}c@{}}Hierarchical \\ Bodies\end{tabular} & \begin{tabular}[c]{@{}c@{}}Hierarchical \\ Headers\end{tabular} & Nested  & Multi-Source & \begin{tabular}[c]{@{}c@{}}Table\\ Language\end{tabular} & \begin{tabular}[c]{@{}c@{}}QA\\ Language\end{tabular} & \begin{tabular}[c]{@{}c@{}}Information\\ Retrieval\end{tabular} & \begin{tabular}[c]{@{}c@{}}Numerical \\ Analysis\end{tabular} & Reasoning & \begin{tabular}[c]{@{}c@{}}Data \\ Analysis\end{tabular} & \begin{tabular}[c]{@{}c@{}}Multi-turn \\ Conversation\end{tabular} & \begin{tabular}[c]{@{}c@{}}Table Structure \\ Understanding\end{tabular} \\ \midrule
WTQ~\cite{pasupat-liang-2015-compositional}        & \ding{51} & \ding{55}                                                        & \ding{55}                                                         & \ding{55} & \ding{55}      & EN                                                       & EN                                                    & \ding{51}                                                         & \ding{51}                                                       & \ding{55}   & \ding{55}                                                  & \ding{55}                                                            & \ding{55}                                                                  \\
SQA~\cite{iyyer-etal-2017-search}        & \ding{51} & \ding{55}                                                        & \ding{55}                                                         & \ding{55} & \ding{55}      & EN                                                       & EN                                                    & \ding{55}                                                         & \ding{55}                                                       & \ding{55}   & \ding{55}                                                  & \ding{51}                                                            & \ding{55}                                                                  \\
TabFact~\cite{Chenwh20TabFact}    & \ding{51}   & \ding{55}                                                        & \ding{55}                                                         & \ding{55} & \ding{55}      & EN                                                       & EN                                                    & \ding{55}                                                         & \ding{55}                                                       & \ding{55}   & \ding{51}                                                  & \ding{55}                                                            & \ding{55}                                                                  \\
FinQA~\cite{chen-etal-2021-finqa}      & \ding{51}   & \ding{55}                                                        & \ding{55}                                                         & \ding{55} & \ding{55}      & EN                                                       & EN                                                    & \ding{55}                                                         & \ding{51}                                                       & \ding{55}   & \ding{55}                                                  & \ding{55}                                                            & \ding{55}                                                                  \\
TAT-QA~\cite{zhu-etal-2021-tat}     & \ding{51}   & \ding{55}                                                        & \ding{55}                                                         & \ding{55} & \ding{55}      & EN                                                       & EN                                                    & \ding{51}                                                         & \ding{51}                                                       & \ding{55}   & \ding{55}                                                  & \ding{55}                                                            & \ding{55}                                                                  \\
OTT-QA~\cite{chen2021open}     & \ding{51}   & \ding{55}                                                        & \ding{55}                                                         & \ding{55} & \ding{55}      & EN                                                       & EN                                                    & -                                                               & -                                                             & \ding{51}   & -                                                        & -                                                                  & -                                                                        \\
FeTaQA~\cite{nan-etal-2022-fetaqa}     & \ding{51}   & \ding{55}                                                        & \ding{55}                                                         & \ding{55} & \ding{55}      & EN                                                       & EN                                                    & \ding{51}                                                         & -                                                             & -         & -                                                        & -                                                                  & -                                                                        \\
AIT-QA~\cite{katsis-etal-2022-ait}     & \ding{51}   & \ding{55}                                                        & \ding{51}                                                         & \ding{55} & \ding{55}      & EN                                                       & EN                                                    & \ding{51}                                                         & \ding{51}                                                       & \ding{55}   & \ding{55}                                                  & \ding{55}                                                            & \ding{55}                                                                  \\
IM-TQA~\cite{zheng-etal-2023-im}     & \ding{51}   & \ding{51}                                                        & \ding{51}                                                         & \ding{55} & \ding{51}      & ZH                                                       & ZH                                                    & \ding{51}                                                         & \ding{55}                                                       & \ding{55}   & \ding{55}                                                  & \ding{55}                                                            & \ding{55}                                                                  \\
TableBench~\cite{wu2024tablebench} & \ding{51}   & \ding{55}                                                        & \ding{55}                                                         & \ding{55} & \ding{51}      & EN                                                       & EN                                                    & \ding{51}                                                         & \ding{51}                                                       & \ding{51}   & \ding{51}                                                  & \ding{55}                                                            & \ding{55}                                                                  \\ \midrule
TableEval (Ours)  & \ding{51}   & \ding{51}                                                        & \ding{51}                                                         & \ding{51} & \ding{51}      & EN/ZH/ZH-HK                                              & ZH                                                    & \ding{51}                                                         & \ding{51}                                                       & \ding{51}   & \ding{51}                                                  & \ding{51}                                                            & \ding{51}                                                                  \\ \bottomrule
\end{tabular}
}
\caption{Comparison of TableEval with other TableQA datasets.}
\label{table-comparison}
\end{table*}

\section{Construction of TableEval} 

\subsection{Task Formulation}
Given a table $T$, which may be presented in complex structures such as hierarchical, nested, or others, and a corresponding natural language question $Q$ related to $T$, the objective is to generate an accurate natural language answer $A$. Unlike existing TableQA tasks that require varied output formats based on the task type, our benchmark standardizes the output to natural language responses to precisely assess the comprehensive TableQA capabilities of LLMs. This approach accommodates various task types within diverse real-world industrial scenarios, including information retrieval, numerical analysis, table reasoning, data analysis, multi-turn dialogue, and understanding of table structures. By supporting complex table structures and diverse data sources, our benchmark ensures that LLMs are evaluated on their ability to understand, interpret, and interact with intricate table data consistently and effectively using natural language.

\subsection{Dataset Construction} \label{construction}
%To comprehensively evaluate table-based question answering under realistic conditions, our dataset construction consists of three steps: (1) Tabular Data Collection, (2) Question Generation, and (3) QA Acquisition and Quality Control. 

\textbf{Tabular Data Collection.} To reduce the risk of data leakage and ensure our dataset accurately reflects current TableQA challenges, we exclusively collect documents published in 2024. We select four document types to capture a broad coverage of industrial and academic contexts: (i) financial reports and company announcements, (ii) industry/stock research reports, (iii) academic papers, and (iv) governmental data. These materials are sourced from publicly available authoritative channels to ensure completeness and legal compliance. Specifically, we collect financial documents and company announcements from the Shenzhen Stock Exchange\footnote{\url{http://www.szse.cn/disclosure/listed/notice/index.html}}, industry/stock research reports from Eastmoney Securities\footnote{\url{https://data.eastmoney.com/report/}}, academic papers from arXiv\footnote{\url{https://arxiv.org}} and the China National Knowledge Infrastructure\footnote{\url{https://www.cnki.net}}, and governmental data from the National Bureau of Statistics\footnote{\url{https://www.stats.gov.cn/sj/zxfb/}}. During the selection, we maintain a balanced distribution of document types to avoid the dataset being biased towards any single source. 

After collecting PDFs (and a limited number of HTML pages), we use parsing tools designed explicitly for PDF and HTML formats to extract tables into Excel worksheets, yielding 29,241 tables. To keep critical contextual information, we apply heuristic rules (see \autoref{rule} and Algorithm~\ref{alg:table_context_extraction} in Appendix for details) to capture any adjacent text snippets near the tables, such as table captions or explanatory notes, so that the semantic context essential for question answering remains. Subsequently, three graduate students with finance and statistics backgrounds review the Excel files to correct parsing errors and ensure the table structures align accurately with the original PDFs. Each table is converted to Markdown format after reviewing. We provide an experiment on the impact of table format in Appendix~\ref{sec:format}.

Next, annotators assign each table to one of seven categories: vertical, horizontal, matrix, concise, hierarchical, nested, and others. Finally, we sample 617 tables by considering factors such as table size, source, content length, and structural type. This sampling approach ensures a diverse yet balanced tabular dataset for benchmarking TableQA.

\textbf{Question Generation.} Based on the verified tables, we generate questions using two strategies, each designed to produce diverse questions.

Template-Prompted Strategy. We design prompt templates to guide GPT-4o in creating questions for various TableQA tasks, including simple lookups, numerical analysis, table size detection, and multi-hop inferences. The template structure and a curated seed pool of examples help guide GPT-4o to generate high-quality questions specific to different task types. However, these generated questions sometimes mirror the examples in the seed pool, limiting the question diversity.

Role-Prompted Strategy. To address the lack of diversity in questions, we simulate role-based scenarios that focus on different aspects of tabular data, such as those of investors, market analysts, and domain experts. Investors might inquire about liquidity or financial stability, while analysts could be more concerned with product strategies and market trends. To ensure consistency with the task types in the Template-Prompted Strategy, we prompt GPT-4o to categorize each generated question into a predefined task type (e.g., simple lookups, numerical analysis). This approach better reflects actual usage contexts yet finds it challenging to produce highly complex or long-tail queries beyond typical stakeholder interests.

Combining both strategies, we obtain an extensive collection of QA pairs associated with multiple user perspectives and table formats. However, the automatically generated answers are often concise and lack detailed reasoning. Therefore, we only keep the generated questions. Additionally, questions often show an uneven distribution of task types due to two primary factors. First, the structure of the tables imposes inherent limitations on the complexity of the questions that can be generated. Second, while the role-prompted strategy uses role-driven scenarios to create questions, predicting the types of questions in advance remains challenging. For example, flat, horizontally oriented tables typically generate simple fact-based questions, such as ``\textit{What is the price of X?}''. In contrast, hierarchical tables, which can represent more complex relationships, are better suited for supporting multi-step inferential or aggregative questions.

\begin{figure}[t]
\centering 
\setlength{\belowcaptionskip}{-0.6cm}
  \includegraphics[width=\linewidth]{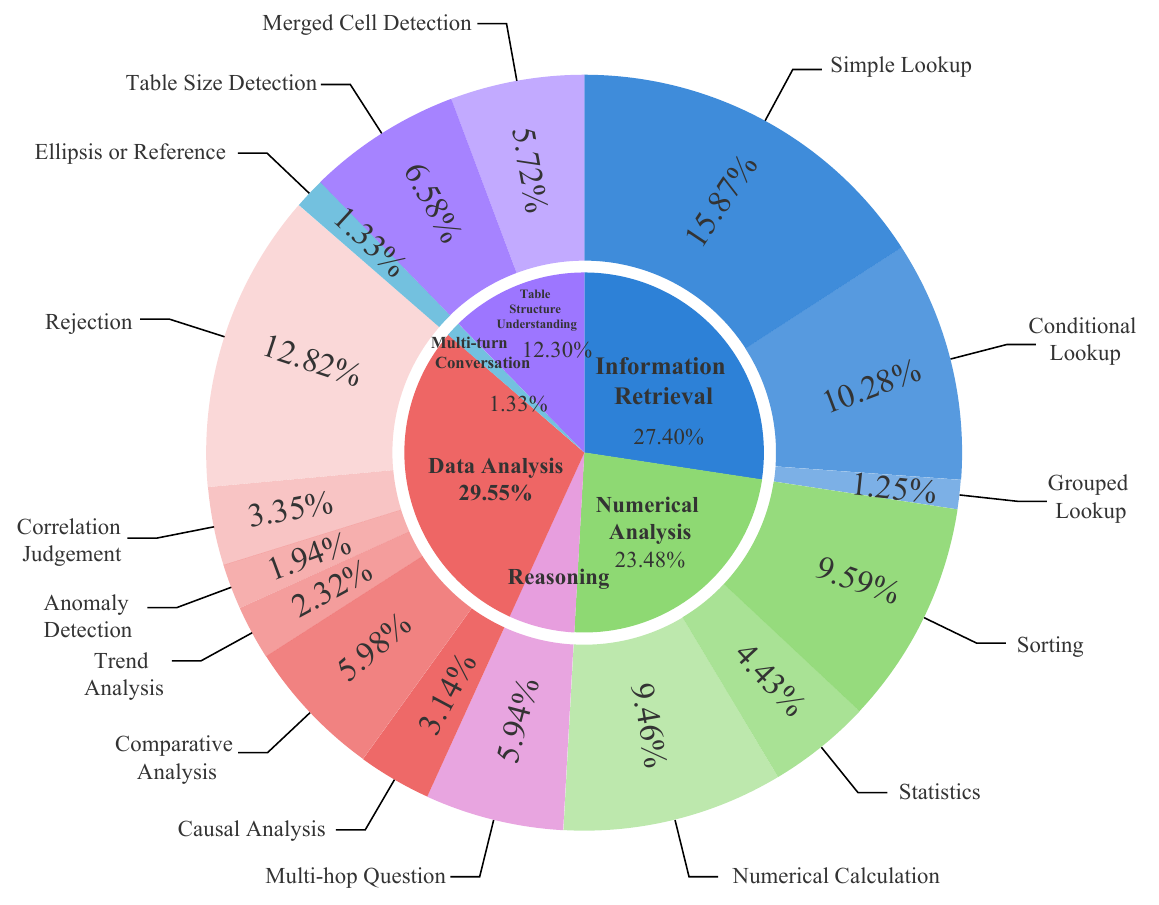}\\
  \caption{Task distribution of our TableEval. \label{task-distribution}}
\end{figure}

We use K-means clustering within each task for sampling and deduplication to enhance question diversity and ensure a balanced task distribution. For each task, we set the expected number of QA pairs as the number of clusters. Specifically, we use BGE-M3~\cite{bge-m3} to obtain question embeddings and cluster them by similarity, retaining only one representative query per cluster. This process yields 34,161 distinct QA pairs. Among these, 5,422 pairs (spanning 16 tasks) are selected as candidates for the final test set, with the remainder serving as the candidate training set.

%We use K-means clustering for sampling and deduplication to enhance question diversity and ensure a balanced task distribution. Specifically, we use BGE-M3~\cite{bge-m3} to vectorize the questions for each task and cluster them by similarity, retaining only one representative query per cluster. This process results in 34,161 distinct questions. Of these, 2,325 (spanning 16 tasks) are candidates for the final dataset.

\textbf{QA Acquisition and Human Annotation.} Human annotation is essential for ensuring accuracy and consistency in our final question-answer pairs. Initially, we employ GPT-4o to generate free-text answers, supplemented by outputs from multiple additional LLMs. Our annotation team consists of six graduate students with finance and statistics backgrounds, trained on comprehensive guidelines covering task categorization, question decomposition, and structured answer formats.

\textbf{First-stage annotation} focuses on quality control and accuracy verification. Annotators review each QA pair according to four principles: eliminating ambiguous or impractical questions, verifying relevance to the underlying table content, correcting inaccuracies in LLM-generated answers, and filtering out overly simple questions (those correctly answered by all LLMs with brief responses). Two independent annotators review each pair, with team discussions to resolve any inconsistencies, followed by expert sampling reviews. If acceptance rates drop below 90\% during sampling review, we initiate additional human verification.

\begin{table}[t]
\small
\centering
\setlength{\belowcaptionskip}{-0.6cm}
\begin{tabular}{@{}cccccccc@{}}
\toprule
\multirow{2}{*}{Type} & \multicolumn{3}{c}{Flat} & \multirow{2}{*}{Hier.} & \multirow{2}{*}{Conc.} & \multirow{2}{*}{Nested} & \multirow{2}{*}{Oth.} \\ \cmidrule(lr){2-4}
                      & V       & H     & M      &                        &                        &                         &                       \\ \midrule
\#Table               & 166     & 6     & 69     & 243                    & 99                    & 13                      & 21                    \\ \bottomrule
\end{tabular}
\caption{Table Structure-wise Statistics. V, H, M, Hier, Conc, and Nested represent vertical, horizontal, matrix, hierarchical, concise, and nested table types.}
\label{table_type}
\end{table}

\begin{figure*}[t]
  \centering 
  \setlength{\belowcaptionskip}{-0.3cm}
  \includegraphics[width=1.0 \linewidth]{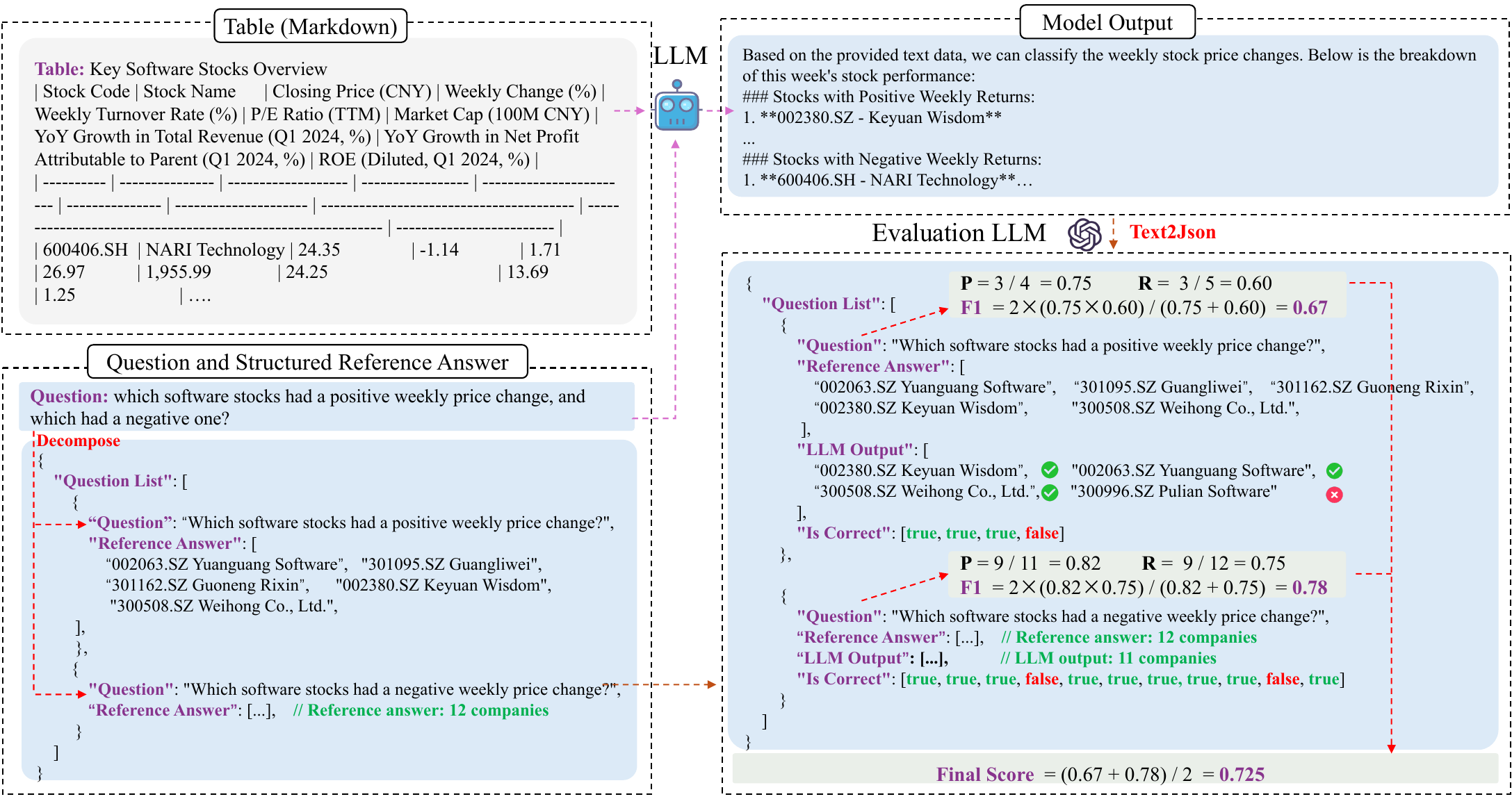}\\
  \caption{Overview of Our SEAT evaluation method. \label{seat}}
\end{figure*}

\textbf{Second-stage annotation} employs a structured answer extraction process that determines whether original questions require decomposition into sub-questions and extracts JSON-formatted answers from annotated responses. This stage filters redundant and low-relevance information from natural language responses, refining answer content to improve comparison accuracy. The JSON-formatted answers undergo two rounds of human verification and one round of expert review to ensure accuracy and consistency with the original table data and question context. The comprehensive annotation process averaged approximately \$2.5 per sample.

\textbf{Second-stage annotation} employs a structured answer extraction process. We use LLMs to determine if the original question needs to be decomposed into sub-questions and to extract JSON-formatted answers from the annotated responses. This process filters redundant and low-relevance information from natural language responses, refining the answer content to improve comparison accuracy. The resulting JSON-formatted answers then undergo two rounds of human verification and one round of expert review to ensure accuracy and consistency with the original table data and question context. The comprehensive annotation process averaged approximately \$2.5 per sample.

% that determines whether original questions require decomposition into sub-questions and extracts JSON-formatted answers from annotated responses. This stage filters redundant and low-relevance information from natural language responses, refining answer content to improve comparison accuracy. The JSON-formatted answers undergo two rounds of human verification and one round of expert review to ensure accuracy and consistency with the original table data and question context. The comprehensive annotation process averaged approximately \$2.5 per sample.

\subsection{Dataset Statistics}

We present the task and table structure distributions of our dataset in \autoref{task-distribution} and \autoref{table_type}, respectively. Our dataset comprises six main task categories, further divided into 16 subcategories, covering real-world scenarios such as simple and conversational retrieval, reasoning, numerical and data analysis, and table structure understanding. It includes 2,325 QA pairs, spanning Simplified Chinese, Traditional Chinese (Hong Kong), and English tables. See more details in \autoref{table-task-stat} and \autoref{tqa_lang}.

Our dataset provides comprehensive coverage of various domain-specific subtasks, supporting a more fine-grained evaluation of cross-domain capabilities. For financial reports, we include subtasks such as financial metric retrieval, ratio and indicator trend analysis, and cross-period comparisons. Industry research reports encompass subtasks like industry performance comparison and financial indicator ranking. Academic papers contain experimental result queries and definition lookups. Government data includes price change calculations and industry growth rate queries.

%\textbf{Data Collection.}
%\textbf{Table Collection.}
%\textbf{Question Construction.}We present the task, table structure, and language distributions of our dataset in \autoref{task-distribution}, \autoref{table_type}, and \autoref{tqa_lang}, respectively. Our dataset comprises six main task categories, further divided into 16 subcategories, covering real-world scenarios such as simple and conversational retrieval, reasoning, numerical and data analysis, and table structure understanding. It includes 2,325 QA pairs, spanning Simplified Chinese, Traditional Chinese (Hong Kong), and English tables. See more details in \autoref{table-task-stat}.
%\textbf{Answer Annotation.}
%\textbf{Quality Control.}
%\textbf{Dataset statistics.}

\section{Evaluation Method for TableQA}

%\begin{figure}[t]
%  \centering 
%  \includegraphics[width=3 in]{imgs/evaluation.pdf}\\
%  \caption{Overview of our proposed evaluation method for TableQA. \label{eval}}
%\end{figure}

Evaluating LLMs in TableQA is challenging because natural language responses can vary widely in phrasing and structure. Existing methods often use exact match~\cite{zhu-etal-2021-tat}, F1-score~\cite{chen-etal-2021-finqa}, or n-gram-matching~\cite{parikh-etal-2020-totto, nan-etal-2022-fetaqa} metrics. However, these approaches often fail to assess the accuracy of LLM-generated responses effectively. To solve this problem, we propose a simple and effective evaluation framework, SEAT (Structured Evaluation for Answers in TableQA), that leverages an LLM with crafted prompts to compare generated responses with structured reference answers.

As shown in \autoref{seat}, SEAT follows a structured process to evaluate the accuracy and relevance of model-generated responses, guided by the original question and a corresponding JSON-formatted reference answer. SEAT consists of two steps: (1) Extract key answers to each sub-question from the model's final response, compare each with the reference answers to identify matching or divergent elements, and label each extracted answer as``true'' or ``false'' based on its alignment with the reference. (2) Present the evaluation results in a JSON format, including the original question, the model’s response, and the evaluation record for each answer to facilitate tracking and verification. Compute the overall F1 for each QA pair by averaging the F1 scores across all sub-questions. For multi-turn conversational QA, first compute the average F1 for each turn, then average across all turns to obtain the final F1 score for the QA session.

Specifically, given a model-generated answer $A$ and a reference answer $R$, both in natural language, SEAT converts them into structured forms and evaluates their alignment in a single step. It uses an LLM guided by a specific prompt (see \autoref{evaluation_prompt}), which can be formalized as:
\begin{equation}
\small
\text{J}_{\text{eval}} = \text{Text2Json}(A, R)
\end{equation}
where Text2Json instructs the LLM to parse both $A$ and $R$. The LLM is prompted to first extract key answer components from $A$ corresponding to the question, then compare these components against the information in $R$ to determine their correctness (labeling each as "true" or "false"). The final output, $\text{J}_{\text{eval}}$, is a single JSON object that contains the structured representations of both answers alongside the correctness judgments for each sub-question. Precision, recall, and the F1 score are then calculated from this structured output. This integrated approach ensures that the semantic comparison is directly tied to the structuring process, providing a robust and consistent evaluation.

%Specifically, given a model-generated answer $A$ and a reference answer $R$, SEAT first converts them into structured forms using a Text2Json operation with a specific prompt (\autoref{evaluation_prompt}) based on an LLM, which can be formalized as:

%\begin{equation}
%\small
%\setlength{\abovedisplayskip}{3pt} % 设置公式上方的间距
%\setlength{\belowdisplayskip}{3pt} % 设置公式下方的间距
%J_A=\textrm{Text2Json}(A), J_R=\textrm{Text2Json}(R)
%\end{equation}
%where $J_A$ and $J_R$ are JSON formats that capture relevant answer components. If the reference answer is not already structured, SEAT applies Text2Json to extract a structured form. 

%Given a question $Q$ and the set of its sub-questions $S=\{S_1, S_2,..., S_n\}$. For each sub-question $S_i$, we use LLM to verify whether $J_A$ correctly reflects the corresponding reference answer $r_i$ using LLM-as-a-judge~\cite{zheng2023judging}:
%\begin{equation}
%\small
%\text{Match}(s_i) = 
%\begin{cases} 
%1, & \text{if } \exists a_j \in J_A \text{, } \text{LLM-as-a-judge}(a_j) = r_i \\
%0, & \text{otherwise}
%\end{cases}
%\end{equation}
%
%The results are recorded as a list, from which precision, recall, and F1 score will be calculated.

\section{Experiment}

%\begin{figure}[t]
%\centering 
%\setlength{\abovecaptionskip}{0.12cm}    
%\setlength{\belowcaptionskip}{-0.6cm}
%  \includegraphics[width=0.8\linewidth]{imgs/lang_and_domain.pdf}\\
%  \caption{Performance of LLMs across languages (a) and domains (b). \label{lang_and_domain}}
%\end{figure}

\subsection{Experimental Setup}
We evaluate 19 models, ranging from 7B to 671B parameters, to assess their performance on TableEval. The evaluated models include open-source and closed-source models to comprehensively compare different model architectures and capabilities. For open-source models, we evaluate Qwen2.5-Instruct series~\cite{qwen25technicalreport}: Qwen2.5-7B-Instruct, Qwen2.5-14B-Instruct, Qwen2.5-32B-Instruct, Qwen2.5-72B-Instruct; Qwen2.5-Coder-Instruct-32B~\cite{hui2024qwen25codertechnicalreport}; QwQ-32B-Preview; DeepSeek series~\cite{deepseekai2024deepseekv2}: DeepSeek-V2-Lite-Chat (16B), DeepSeek-Coder-V2-Lite-Instruct (16B), DeepSeek-V2.5-1210 (236B), DeepSeek-V3 (671B), DeepSeek-R1 (671B)~\cite{deepseekr1}; glm-4-9b-Chat (9B)~\cite{glm2024}; Llama-3 series~\cite{llama3}: Llama-3.1-8B-Instruct, Llama-3.3-70B-Instruct. For closed-source models, we evaluate qwen-max-2024-09-19, Claude 3.5 sonnet-20241022, GPT-4o-mini-2024-07-18, GPT-4o-2024-11-20~\cite{gpt4}, and o1-preview. We set the temperature to 0 and use a fixed seed (42) to make the model's output more deterministic. For models that do not allow the configuration of parameters (e.g., temperature or seed), we use their default settings.

\begin{table}[t]
\centering
\setlength{\belowcaptionskip}{-0.3cm}
\resizebox{\linewidth}{!}{
\begin{tabular}{@{}llccc@{}}
\toprule
\multicolumn{2}{l}{Metric}                      & $r$            & $\rho$         & $\tau$         \\ \midrule
\multicolumn{2}{l}{F1}                          & .1776          & .1222          & .1014          \\
\multicolumn{2}{l}{ROUGE-L}                     & .6132          & .6030          & .4988          \\
\multirow{3}{*}{LLM-as-a-judge} & GPT4o         & .8673          & .8792          & .7733          \\
                                & Qwen-2.5-72B  & .7080          & .7333          & .6592          \\
                                & Llama-3.3-70B & .8052          & .8311          & .7214          \\ \midrule
\multirow{3}{*}{SEAT}           & GPT4o         & \textbf{.9373} & \textbf{.9346} & \textbf{.9062} \\
                                & Qwen-2.5-72B  & .8257          & .8137          & .7726          \\
                                & Llama-3.3-70B & .8410          & .8359          & .8214          \\ \bottomrule
\end{tabular}
}
\caption{Correlation with human scores, measured with Pearson $r$, Spearman $\rho$, and Kendall $\tau$ coefficients. We evaluate SEAT with different LLM backbones. LLM-as-a-judge~\cite{zheng2023judging} prompts GPT-4o to provide a 1-10 rating for evaluation.}
\label{correlation}
\end{table}

\begin{table*}[t]
\centering
\small
\setlength{\belowcaptionskip}{-0.3cm}
\resizebox{\linewidth}{!}{
\begin{tabular}{@{}lcccccccc@{}}
\toprule
\textbf{Model}                  & \textbf{\begin{tabular}[c]{@{}c@{}}Size\\ (B)\end{tabular}} & \textbf{Avg}   & \textbf{\begin{tabular}[c]{@{}c@{}}Information \\ Retrieval\end{tabular}} & \textbf{\begin{tabular}[c]{@{}c@{}}Numerical \\ Analysis\end{tabular}} & \textbf{Reasoning} & \textbf{\begin{tabular}[c]{@{}c@{}}Data \\ Analysis\end{tabular}} & \textbf{\begin{tabular}[c]{@{}c@{}}Multi-turn \\ Conversation\end{tabular}} & \textbf{\begin{tabular}[c]{@{}c@{}}Table Structure \\ Understanding\end{tabular}} \\ \midrule
Qwen2.5-7B-Instruct             & 7                                                           & 59.60          & 69.23                                                                     & 64.29                                                                  & 59.38              & 69.71                                                             & 68.67                                                                       & 26.35                                                                             \\
Llama-3.1-8B-Instruct           & 8                                                           & 49.26          & 67.40                                                                     & 53.35                                                                  & 48.82              & 57.06                                                             & 53.15                                                                       & 15.76                                                                             \\
glm-4-9b-chat                   & 9                                                           & 53.61          & 66.19                                                                     & 51.09                                                                  & 55.09              & 62.47                                                             & 64.36                                                                       & 22.44                                                                             \\
Qwen2.5-14B-Instruct            & 14                                                          & 70.02          & 84.72                                                                     & 78.93                                                                  & 68.65              & 75.06                                                             & 75.05                                                                       & 37.72                                                                             \\
DeepSeek-V2-Lite-Chat           & 16                                                          & 36.75          & 48.52                                                                     & 35.43                                                                  & 35.97              & 51.80                                                             & 41.61                                                                       & 7.15                                                                              \\
DeepSeek-Coder-V2-Lite-Instruct & 16                                                          & 48.30          & 60.40                                                                     & 56.39                                                                  & 50.03              & 51.51                                                             & 50.62                                                                       & 20.83                                                                             \\
QwQ-32B-Preview                 & 32                                                          & 78.14          & 89.33                                                                     & 85.75                                                                  & 81.37              & 71.69                                                             & 82.15                                                                       & 58.53                                                                             \\
Qwen2.5-32B-Instruct            & 32                                                          & 75.50          & 86.32                                                                     & 84.10                                                                  & 76.09              & 77.60                                                             & 82.25                                                                       & 46.61                                                                             \\
Qwen2.5-Coder-32B-Instruct      & 32                                                          & 70.75          & 79.82                                                                     & 77.00                                                                  & 73.03              & 76.33                                                             & 74.89                                                                       & 43.44                                                                             \\
Llama-3.3-70B-Instruct          & 70                                                          & 72.94          & 87.42                                                                     & 76.70                                                                  & 73.38              & \textbf{81.27}                                                    & 80.62                                                                       & 38.24                                                                             \\
Qwen2.5-72B-Instruct            & 72                                                          & 74.23          & 82.68                                                                     & 81.53                                                                  & 74.85              & 78.94                                                             & 81.90                                                                       & 45.50                                                                             \\
DeepSeek-V2.5-1210                   & 236                                                         & 73.27          & 87.41                                                                     & 79.10                                                                  & 71.49              & 77.97                                                             & 78.72                                                                       & 44.94                                                                             \\
DeepSeek-V3                     & 671                                                         & 77.95          & \textbf{91.20}                                                            & 82.61                                                                  & 81.72              & 77.45                                                             & \textbf{85.83}                                                              & 48.89                                                                             \\
DeepSeek-R1                     & 671                                                         & \textbf{82.46} & 90.15                                                                     & \textbf{88.56}                                                         & \textbf{87.91}     & 77.79                                                             & 78.29                                                                       & \textbf{72.05}                                                                    \\ \midrule
qwen-max-2024-09-19             & N/A                                                         & 73.34          & 84.42                                                                     & 81.35                                                                  & 72.64              & 78.09                                                             & 80.18                                                                       & 43.35                                                                             \\
Claude-3-5-sonnet-20241022      & N/A                                                         & 83.32          & \textbf{89.62}                                                            & \textbf{91.06}                                                         & \textbf{85.76}     & \textbf{84.01}                                                    & \textbf{87.94}                                                              & 61.51                                                                             \\
gpt-4o-mini-2024-07-18          & N/A                                                         & 68.47          & 82.64                                                                     & 76.15                                                                  & 73.13              & 70.70                                                             & 73.66                                                                       & 34.56                                                                             \\
gpt-4o-2024-11-20               & N/A                                                         & 78.79          & 88.24                                                                     & 86.00                                                                  & 83.05              & 81.47                                                             & 83.20                                                                       & 50.79                                                                             \\
o1-preview                      & N/A                                                         & \textbf{83.43} & 88.30                                                                     & 87.08                                                                  & 82.88              & 77.89                                                             & 83.38                                                                       & \textbf{81.03}                                                                    \\ \bottomrule
\end{tabular}
}
\caption{Performance of LLMs on TableEval. We report the F1 score calculated by our SEAT. Bold values indicate the best result within each category. Avg denotes the overall score.}
\label{task-major-type}
\end{table*}

To obtain detailed, interpretable, and structured responses for each question in the TableQA task, we conducted comparative experiments based on Chain-of-Thought (CoT) prompting~\cite{wei2022chain}. The specific CoT prompt used in our experiments are shown in the \autoref{cot_prompt}, which align with those in previous studies~\cite{zheng-etal-2024-multimodal, deng-etal-2024-tables, zhang-etal-2024-e5}.

\subsection{Effectiveness of SEAT}

To assess the effectiveness of SEAT, we analyze the correlation between SEAT and human scores. Annotators rate model responses from 1 to 10 based on reference quality. We collect ratings for 5\% randomly selected samples and evaluate the correlation using three metrics: 1) Pearson correlation coefficient ($r$), 2) Spearman rank coefficient ($\rho$), and 3) Kendall rank coefficient ($\tau$). 
As shown in \autoref{correlation}, SEAT consistently achieves higher correlations with human judgments than existing metrics (F1 and ROUGE-L) and the LLM-as-a-judge. When using GPT-4o as the backbone, SEAT attains the highest agreement with human judgments. This improvement over the direct rating approach (LLM-as-a-judge) remains significant across all tested backbones, including Qwen-2.5-72B and Llama-3.3-70B, demonstrating SEAT’s robustness.

\subsection{Main Results}

\autoref{task-major-type} reports how well each LLM handles six main tasks in TableEval: \emph{Information Retrieval}, \emph{Numerical Analysis}, \emph{Reasoning}, \emph{Data Analysis}, \emph{Multi-turn Conversation}, and \emph{Table Structure Understanding}, along with the average score. Larger models generally perform better, with closed-source models like Claude 3.5 Sonnet, GPT-4o, and o1-preview often leading the rankings. o1-preview achieves the highest average score (83.43\%), while Claude 3.5 Sonnet performs comparably (83.32\%). Both perform well in complex tasks such as reasoning and data analysis, which require deep reasoning and domain knowledge. Among open-source models, DeepSeek-R1 shows strong, complex reasoning abilities and performs best in reasoning, numerical analysis, and table structure understanding. Qwen2.5-14B-Instruct significantly improves over its 7B version, demonstrating the benefits of increased parameters. Scaling up to 32B further enhances overall performance, with models like QwQ-32B-Preview and Qwen2.5-32B-Instruct improving across most metrics. However, even these large models still struggle with table structure understanding, suggesting that specialized training is necessary for tasks requiring layout comprehension. The significant gap between top closed-source LLMs and mid-sized open-source ones underscores the advantages that closed-source models may have, potentially through specialized training strategies, proprietary datasets, or more computational resources.

\begin{figure}[t]
\centering 
\setlength{\belowcaptionskip}{-0.6cm}
  \includegraphics[width=\linewidth]{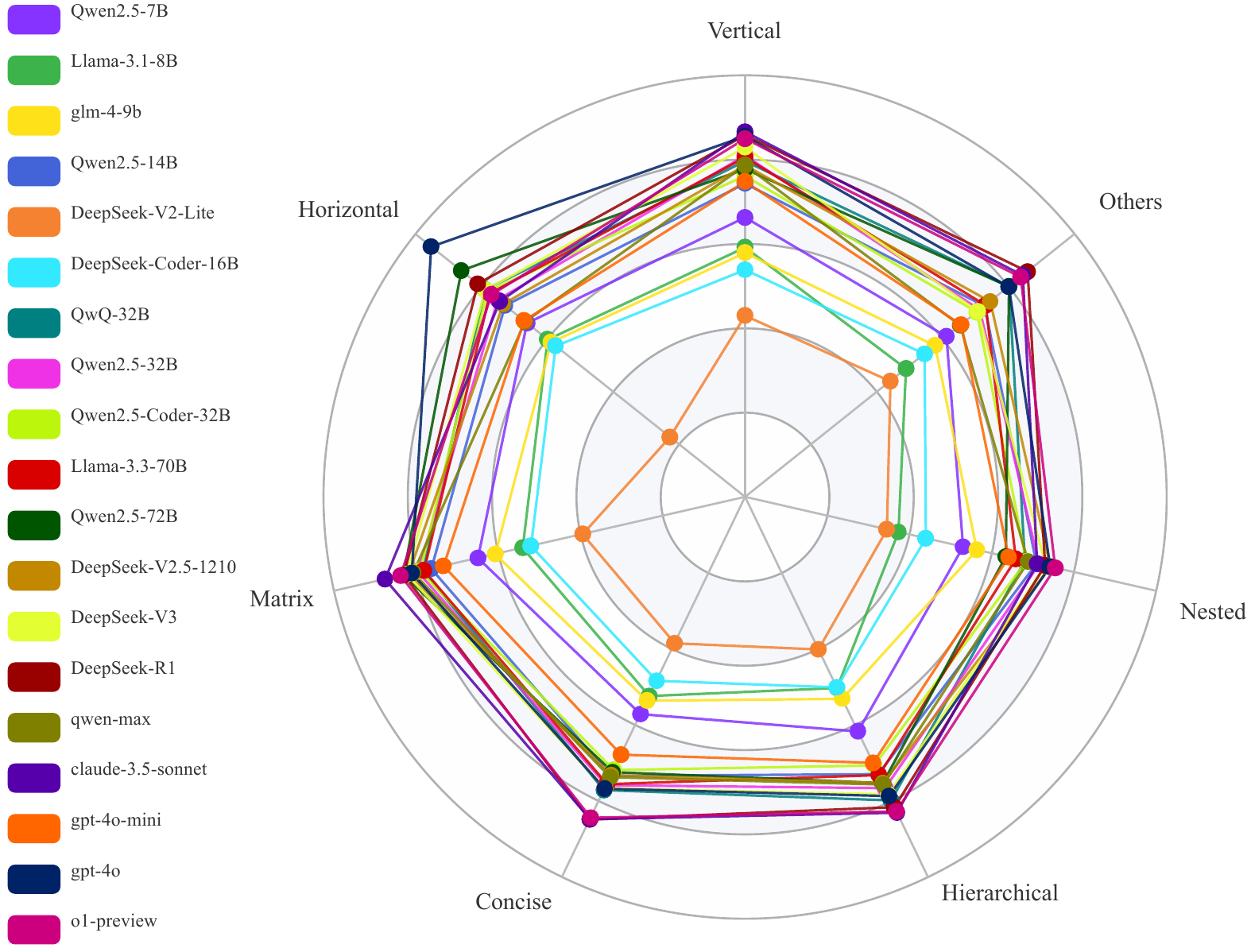}\\
  \caption{Performance of LLMs across table structures. \label{radar}}
\end{figure}

\textbf{Structure-aspect Performance.} \autoref{radar} (See \autoref{structure-level} for details) depicts the performance of different LLMs across different table structures, including \emph{vertical}, \emph{horizontal}, \emph{matrix}, and more complex \emph{concise}, \emph{hierarchical}, and \emph{nested} structures. Closed-source models generally demonstrate more balanced and superior performance. GPT-4o, Claude, and o1-preview outperform across different table structures, demonstrating their strong capabilities. However, specific open-source models can be highly competitive in specific structures. For example, DeepSeek R1 performs particularly well on \textit{hierarchical} and \textit{others} table structures. Besides, all models, whether open or closed, face more significant challenges with \textit{nested} and \textit{hierarchical} tables compared to simpler formats, underscoring the ongoing difficulty in structure-aware comprehension and structured reasoning.

\begin{table}[t]
\tiny
\centering
\setlength{\abovecaptionskip}{0.12cm}    
\setlength{\belowcaptionskip}{-0.6cm}
\resizebox{\linewidth}{!}{
\begin{tabular}{@{}lcccc@{}}
\toprule
Model                           & Size & ZH             & ZH-HK          & EN             \\ \midrule
Qwen2.5-7B-Instruct             & 7    & 61.98          & 61.62          & 63.67          \\
Llama-3.1-8B-Instruct           & 8    & 53.24          & 52.48          & 54.66          \\
glm-4-9b-chat                   & 9    & 55.77          & 55.19          & 55.30          \\
Qwen2.5-14B-Instruct            & 14   & 72.00          & 74.49          & 76.96          \\
DeepSeek-V2-Lite-Chat           & 16   & 40.56          & 43.97          & 38.09          \\
DeepSeek-Coder-V2-Lite-Instruct & 16   & 51.32          & 51.66          & 50.66          \\
QwQ-32B-Preview                 & 32   & 78.55          & 76.40          & 81.62          \\
Qwen2.5-32B-Instruct            & 32   & 76.99          & 78.43          & 78.84          \\
Qwen2.5-Coder-32B-Instruct      & 32   & 73.03          & 70.12          & 75.70          \\
Llama-3.3-70B-Instruct          & 70   & 75.52          & 77.13          & 76.79          \\
Qwen2.5-72B-Instruct            & 72   & 76.16          & 74.51          & 77.68          \\
DeepSeek-V2.5-1210              & 236  & 75.51          & 78.15          & 77.31          \\
DeepSeek-V3                     & 671  & 78.20          & 79.69          & 81.60          \\
DeepSeek-R1                     & 671  & \textbf{83.46} & \textbf{82.27} & \textbf{84.86} \\ \midrule
qwen-max-2024-09-19             & N/A  & 74.94          & 78.36          & 76.88          \\
Claude-3-5-sonnet-20241022      & N/A  & \textbf{84.50} & \textbf{83.84} & 85.34          \\
gpt-4o-mini-2024-07-18          & N/A  & 69.64          & 73.80          & 72.17          \\
gpt-4o-2024-11-20               & N/A  & 80.14          & 79.73          & 82.75          \\
o1-preview                      & N/A  & 82.69          & 82.85          & \textbf{86.41} \\ \bottomrule
\end{tabular}
}
\caption{Language-aspect performance.}
\label{lang-level}
\end{table}

\textbf{Language-aspect Performance.} \autoref{lang-level} depicts the performance of different LLMs across tables in Simplified Chinese, Traditional Chinese, and English. Closed-source models demonstrate strong multilingual capabilities, maintaining high and consistent accuracy across these languages. Claude 3.5 Sonnet and o1-preview consistently achieve the top performance. A notable trend is a model's development background's influence on language strengths. Models trained in large Chinese corpora, such as the Qwen series and DeepSeek-R1, tend to perform best in English, followed by Simplified Chinese, with Traditional Chinese ranking lowest. In contrast, models primarily trained in English often achieve their highest accuracy in English or Traditional Chinese, while Simplified Chinese tends to perform worse.

\textbf{Domain-aspect Performance.} \autoref{domain-level} shows the performance of different LLMs across tables in finance, government, paper, and report domains. Larger models consistently outperform smaller ones across all four domains. Smaller models show more significant performance variation, struggling in specific areas, whereas larger models remain more stable, indicating more substantial generalization across diverse tabular data. The results show that Finance and Government data are more challenging, as smaller models tend to have greater performance fluctuations. On the other hand, Paper and Report tasks are handled more steadily, suggesting structured data complexity varies by field.

\textbf{Key Insights and Observations.} (1) Closed-Source Models Lead, but Open-Source Models Show Promise. Closed-source models (e.g., Claude 3.5 Sonnet, GPT-4o, o1-preview) consistently achieve top-tier performance across tasks, table structures, languages, and domains, benefiting from specialized training resources. However, large-scale open-source models (e.g., DeepSeek-R1, QwQ-32B-Preview) can be highly competitive with test-time scaling, indicating that open models have the potential to reduce the gap on complex TableQA with continued improvements. (2) Table Structure Understanding Remains a Challenge. Both closed- and open-source models struggle with complex table structures, particularly nested and hierarchical structures. Performance drops of 10–15\% are observed compared to simple structures (flat tables), highlighting the importance of future research on explicit table structures or structure-aware encoding methods. (3) Domain and Language-Specific Performance Gaps. Open-source models like DeepSeek-R1 perform strongly in specific domains, such as government data, while closed-source models demonstrate strong cross-domain adaptability. Similarly, language-aspect evaluation shows that while models often perform well in English, their performance declines in Simplified or Traditional Chinese. These findings suggest that domain-specific of language-specific training strategies should be explored. 

\begin{table}[t]
\tiny
\centering
\setlength{\abovecaptionskip}{0.12cm}    
\setlength{\belowcaptionskip}{-0.6cm}
\resizebox{\linewidth}{!}{
\begin{tabular}{@{}lccccc@{}}
\toprule
Model                           & Size & Finance & Government & Paper & Report \\ \midrule
Qwen2.5-7B-Instruct             & 7    & 62.69   & 60.44      & 63.60 & 60.94  \\
Llama-3.1-8B-Instruct           & 8    & 53.47   & 49.04      & 55.87 & 52.63  \\
glm-4-9b-chat                   & 9    & 55.49   & 54.42      & 54.94 & 57.10  \\
Qwen2.5-14B-Instruct            & 14   & 73.11   & 73.32      & 74.90 & 73.55  \\
DeepSeek-V2-Lite-Chat           & 16   & 43.02   & 34.84      & 38.71 & 40.57  \\
DeepSeek-Coder-V2-Lite-Instruct & 16   & 50.88   & 46.14      & 50.76 & 55.23  \\
QwQ-32B-Preview                 & 32   & 77.71   & 79.24      & 80.09 & 79.97  \\
Qwen2.5-32B-Instruct            & 32   & 77.30   & 74.56      & 79.68 & 77.64  \\
Qwen2.5-Coder-32B-Instruct      & 32   & 71.65   & 72.28      & 76.19 & 73.06  \\
Llama-3.3-70B-Instruct          & 70   & 76.43   & 70.77      & 76.93 & 77.23  \\
Qwen2.5-72B-Instruct            & 72   & 73.71   & 76.20      & 78.90 & 78.32  \\
DeepSeek-V2.5-1210              & 236  & 77.31   & 74.85      & 75.30 & 76.86  \\
DeepSeek-V3                     & 671  & 79.03   & 79.73      & 79.45 & 79.57  \\
DeepSeek-R1                     & 671  & 81.30   & 90.13      & 84.86 & 83.48  \\ \midrule
qwen-max-2024-09-19             & N/A  & 76.19   & 75.57      & 75.78 & 76.27  \\
Claude-3-5-sonnet-20241022      & N/A  & 82.75   & 86.18      & 86.61 & 85.08  \\
gpt-4o-mini-2024-07-18          & N/A  & 71.34   & 68.37      & 70.79 & 71.96  \\
gpt-4o-2024-11-20               & N/A  & 79.63   & 81.29      & 83.18 & 79.57  \\
o1-preview                      & N/A  & 82.85   & 86.36      & 84.34 & 83.06  \\ \bottomrule
\end{tabular}
}
\caption{Domain-aspect performance.}
\label{domain-level}
\end{table}

\section{Conclusion}

We introduce TableEval, a real-world benchmark that fills a critical gap in evaluating LLMs on TableQA by featuring various table structures, languages, and domains. By curating recently published tables, TableEval minimizes data leakage and provides a more accurate evaluation of LLMs. Experimental results show that while closed-source models generally perform better, well-scaled and fine-tuned open-source models remain competitive. However, both struggle with complex table structures like nested or hierarchical ones, leading to a 10–15\% performance drop. Additionally, models still face challenges with domain-specific and cross-lingual understanding. Besides, our SEAT evaluation method assesses answers structurally, ensuring semantic accuracy where traditional exact-match and n-gram matching metrics fall short and achieving high consistency with human judgments. We believe TableEval and SEAT will advance research in structure-aware approaches, multilingual representation, and domain adaptation for TableQA.

\section{Limitations}

We conclude the limitations of our study as follows: (1) Despite efforts to source heterogeneous tables (including concise, hierarchical, and nested structures), real-world data includes many other layouts (such as highly irregular formats or image-based tables), that are not fully captured in our benchmark. (2) Our dataset primarily presents questions in Simplified Chinese, while we include table content in Simplified Chinese, Traditional Chinese, and English. Deeper investigations into fully multilingual TableQA scenarios are still needed.

\section*{Acknowledgements}

This work is supported by Beijing Municipal Science \& Technology Commission, Administrative Commission of Zhongguancun Science Park No.Z231100007423016, the National Natural Science Foundation of China under Grants No.62206287. We sincerely thank the anonymous reviewers for their insightful comments and constructive suggestions.

%The research work described in this paper has been supported by the National Natural Science Foundation of China under Grants No. 62206287. We sincerely thank the anonymous reviewers for their insightful comments and constructive suggestions.

%\textbf{Citation Generation.} 
%
%\textbf{Grounded Answer Generation.}
%
%\textbf{Attributable to Identified Sources.}
%
%\textbf{Evidence Retrieval.}

\bibliography{custom}

\appendix

\section{Dataset Construction Details}
\label{sec:appendix}

\begin{table}[h]
\centering
\setlength{\abovecaptionskip}{0.12cm}    
\setlength{\belowcaptionskip}{-0.3cm}
\small
\begin{tabular}{@{}ccc@{}}
\toprule
\textbf{Task Type}                                                                                             & \textbf{Question Category}                 & \textbf{Count} \\ \midrule
\multicolumn{1}{c|}{\multirow{3}{*}{\begin{tabular}[c]{@{}c@{}}Information \\ Retrieval\end{tabular}}}         & \multicolumn{1}{c|}{Simple Lookup}         & 369            \\ \cmidrule(l){2-3} 
\multicolumn{1}{c|}{}                                                                                          & \multicolumn{1}{c|}{Conditional Lookup}    & 239            \\ \cmidrule(l){2-3} 
\multicolumn{1}{c|}{}                                                                                          & \multicolumn{1}{c|}{Grouped Lookup}        & 29             \\ \midrule
\multicolumn{1}{c|}{\multirow{3}{*}{\begin{tabular}[c]{@{}c@{}}Numerical \\ Analysis\end{tabular}}}            & \multicolumn{1}{c|}{Sorting}               & 223            \\ \cmidrule(l){2-3} 
\multicolumn{1}{c|}{}                                                                                          & \multicolumn{1}{c|}{Statistics}            & 103            \\ \cmidrule(l){2-3} 
\multicolumn{1}{c|}{}                                                                                          & \multicolumn{1}{c|}{Numerical Calculation} & 220            \\ \midrule
\multicolumn{1}{c|}{Reasoning}                                                                                 & \multicolumn{1}{c|}{Multi-hop Question}    & 138            \\ \midrule
\multicolumn{1}{c|}{\multirow{6}{*}{\begin{tabular}[c]{@{}c@{}}Data \\ Analysis\end{tabular}}}                 & \multicolumn{1}{c|}{Causal Analysis}       & 73             \\ \cmidrule(l){2-3} 
\multicolumn{1}{c|}{}                                                                                          & \multicolumn{1}{c|}{Comparative Analysis}  & 139            \\ \cmidrule(l){2-3} 
\multicolumn{1}{c|}{}                                                                                          & \multicolumn{1}{c|}{Trend Analysis}        & 54             \\ \cmidrule(l){2-3} 
\multicolumn{1}{c|}{}                                                                                          & \multicolumn{1}{c|}{Anomaly Detection}     & 45             \\ \cmidrule(l){2-3} 
\multicolumn{1}{c|}{}                                                                                          & \multicolumn{1}{c|}{Correlation Judgement} & 78             \\ \cmidrule(l){2-3} 
\multicolumn{1}{c|}{}                                                                                          & \multicolumn{1}{c|}{Rejection}             & 298            \\ \midrule
\multicolumn{1}{c|}{\begin{tabular}[c]{@{}c@{}}Multi-turn \\ Dialogue\end{tabular}}                            & \multicolumn{1}{c|}{Ellipsis or Reference} & 31             \\ \midrule
\multicolumn{1}{c|}{\multirow{2}{*}{\begin{tabular}[c]{@{}c@{}}Table Structure \\ Understanding\end{tabular}}} & \multicolumn{1}{c|}{Table Size Detection}  & 153            \\ \cmidrule(l){2-3} 
\multicolumn{1}{c|}{}                                                                                          & \multicolumn{1}{c|}{Merged Cell Detection} & 133            \\ \midrule
\multicolumn{2}{c|}{Total}                                                                                                                                  & 2,325          \\ \bottomrule
\end{tabular}
\caption{Task Types and Question Categories}
\label{table-task-stat}
\end{table}

\begin{table}[h]
\small
\centering
\setlength{\abovecaptionskip}{0.12cm}    
\setlength{\belowcaptionskip}{-0.6cm}
\begin{tabular}{@{}ccc@{}}
\toprule
Language & \multicolumn{1}{l}{\#Table} & \multicolumn{1}{l}{\#QA Pairs} \\ \midrule
ZH-CN    & 364                         & 1,331                          \\
ZH-HK    & 114                         & 406                            \\
EN       & 139                         & 588                          \\ \midrule
Total    & 617                         & 2,325                          \\ \bottomrule
\end{tabular}
\caption{TableEval Language-wise Statistics.}
\label{tqa_lang}
\end{table}

\subsection{Multi-answer Consistency Checking}
To enhance the quality of data annotation, we leverage GPT-4o, DeepSeek 2.5, and Qwen 2.5-72B-Instruct to generate annotated answers for all questions. A consistency check is then performed on these annotations, with the results serving as guidance for subsequent human check. To this end, we propose an LLM-based consistency-checking method that rigorously evaluates the alignment of answer content (prompt can be found in \autoref{consistency}). This method assesses semantic matching, expression style, core information, and the complete consistency of table content. Any semantic conflicts, redundant details, or incomplete matches identified in the answers are considered inconsistencies during the evaluation process.

\textbf{Semantic Analysis}. The core information in all three answers is compared individually to ensure there are no semantic differences in their expressions. Variations in phrasing, wording, or inconsistencies in described details will make the answers considered inconsistent.

\textbf{Redundancy Check}. If any answer contains unnecessary explanations or repeated content that unnecessarily increases the length, it will also be considered inconsistent.

%\begin{figure*}[t]
%  \centering 
%  \setlength{\abovecaptionskip}{0.2cm}    
%  \setlength{\belowcaptionskip}{-0.2cm}
%  \includegraphics[width=0.92 \linewidth]{imgs/seat.pdf}\\
%  \caption{Comparison of Our SEAT Evaluation Method with Existing LLM-as-a-Judge Methods. \label{seat}}
%\end{figure*}

\begin{table}[t]
\small
\setlength{\abovecaptionskip}{0.12cm}    
\setlength{\belowcaptionskip}{-0.5cm}
\begin{tabular}{@{}lccc@{}}
\toprule
Task                          & $r$    & $\rho$ & $\tau$ \\ \midrule
Information Retrieval         & 0.9687 & 0.9141 & 0.9046 \\
Numerical Analysis            & 0.8509 & 0.8330 & 0.8163 \\
Reasoning                     & 1.0000 & 1.0000 & 1.0000 \\
Data Analysis                 & 0.8972 & 0.8871 & 0.8361 \\
Multi-turn Conversation       & 1.0000 & 1.0000 & 1.0000 \\
Table Structure Understanding & 0.9957 & 0.9957 & 0.9951 \\ \bottomrule
\end{tabular}
\caption{Task-specific correlation of SEAT with human scores, measured with Pearson $r$, Spearman $\rho$, and Kendall $\tau$ coefficients, using GPT-4o as backbone.}
\label{correlation-task}
\end{table}

\subsection{Structured Answer Extraction}

We use GPT-4o to extract structured answers from LLM-generated responses. This method aims to precisely extract the content corresponding to the question from the answer text, which is then used for subsequent model evaluation. It follows the following rules to ensure the accuracy of the extracted answers in terms of values, format, and content. We show the prompt in \autoref{structured}.

\textbf{Multi-part Question Processing}. The original question should be decomposed into a list of sub-questions when it contains multiple sub-questions. 

\textbf{Single Question Processing}. For a single question, keep the answer to that question without unnecessary decomposition, ensuring a concise and clear response.

\textbf{Answer Extraction}. Our goal is to extract concise final answers for each sub-question to ensure an accurate evaluation of all models' natural language responses. Additionally, we have two specific requirements: (a) During the answer extraction process, the original format of numerical values, including percentage signs, units, commas, and scientific notation, must be strictly kept, ensuring the precision of the extracted answers. (b) If the answer text contains ambiguity, errors, or incomplete information, the answer will be labeled as ``Reference Answer Incorrect'', helping the user understand the potential uncertainty of the answer.

\section{Additional Experimental Analysis}
\subsection{Robustness of SEAT}

To provide a more detailed analysis of SEAT's effectiveness across different contexts, we further categorize our main experimental results (\autoref{correlation}) by task type and domain. This analysis aims to determine whether SEAT maintains consistent correlation with human judgments regardless of the specific nature of the TableQA task or the domain from which the tables originate. Rather than conducting separate experiments, we classify our existing evaluation data according to six distinct task types and four domains to examine SEAT's robustness across these dimensions.

\textbf{Task-Specific Robustness.} \autoref{correlation-task} presents the correlation between SEAT and human judgments across six task types in TableEval, using GPT-4o as the evaluation backbone. The results demonstrate SEAT's high consistency across varied task complexities. For reasoning and multi-turn conversation tasks, SEAT achieves perfect correlation with human judgments, indicating that the structured evaluation approach is particularly effective for complex, multi-step reasoning processes. Information retrieval and table structure understanding tasks also show very high correlations (Pearson values of 0.9687 and 0.9957 respectively), suggesting SEAT accurately captures the quality of responses for these fundamental TableQA capabilities.

\begin{table}[t]
\small
\setlength{\abovecaptionskip}{0.12cm}    
\setlength{\belowcaptionskip}{-0.6cm}
\begin{tabular}{@{}lccc@{}}
\toprule
Domain                                                                                 & $r$    & $\rho$ & $\tau$ \\ \midrule
\begin{tabular}[c]{@{}l@{}}financial reports and \\ company announcements\end{tabular} & 0.9575 & 0.9295 & 0.8967 \\
industry/stock research reports                                                        & 0.9605 & 0.9642 & 0.9516 \\
academic papers                                                                        & 0.8794 & 0.8777 & 0.8470 \\
governmental data                                                                      & 0.9973 & 0.9982 & 0.9962 \\ \bottomrule
\end{tabular}
\caption{Domain-specific correlation of SEAT with human scores, measured with Pearson $r$, Spearman $\rho$, and Kendall $\tau$ coefficients, using GPT-4o as backbone.}
\label{correlation-domain}
\end{table}

\begin{table*}[t]
\centering
\small
\setlength{\abovecaptionskip}{0.12cm}    
\setlength{\belowcaptionskip}{-0.2cm}
 \resizebox{\linewidth}{!}{
\begin{tabular}{@{}lcccccccccccccccccc@{}}
\toprule
\multirow{2}{*}{\textbf{Model}} & \multirow{2}{*}{\textbf{Size}} & \multicolumn{1}{c|}{\multirow{2}{*}{\textbf{AVG}}} & \multicolumn{3}{c|}{\textbf{Information Retrival}}             & \multicolumn{3}{c|}{\textbf{Numerical Analysis}}                   & \multicolumn{1}{c|}{\textbf{Reasoning}} & \multicolumn{6}{c|}{\textbf{Data Analysis}}                                                                                                                                                                                                                                                                                                                 & \multicolumn{1}{c|}{\textbf{Conversation}}                                         & \multicolumn{2}{c}{\textbf{Table Structure Understanding}}                                                                         \\ \cmidrule(l){4-19} 
                                &                                & \multicolumn{1}{c|}{}                              & Simple         & Condition      & \multicolumn{1}{c|}{Grouped} & Sorting        & Statistics     & \multicolumn{1}{c|}{Calculation} & \multicolumn{1}{c|}{Multi-hop}          & \begin{tabular}[c]{@{}c@{}}Causal \\ Analysis\end{tabular} & \begin{tabular}[c]{@{}c@{}}Comparative \\ Analysis\end{tabular} & \begin{tabular}[c]{@{}c@{}}Trend \\ Analysis\end{tabular} & \begin{tabular}[c]{@{}c@{}}Anomaly \\ Detection\end{tabular} & \begin{tabular}[c]{@{}c@{}}Correlation \\ Judgement\end{tabular} & \multicolumn{1}{c|}{Rejection} & \multicolumn{1}{c|}{\begin{tabular}[c]{@{}c@{}}Ellipsis/\\ Reference\end{tabular}} & \begin{tabular}[c]{@{}c@{}}Table Size \\ Detection\end{tabular} & \begin{tabular}[c]{@{}c@{}}Merged Cell \\ Detection\end{tabular} \\ \midrule
Qwen2.5-7B-Instruct             & 7                              & 59.94                                              & 69.79          & 68.88          & 65.03                        & 74.57          & 57.28          & 57.14                            & 59.38                                   & 54.25                                                      & 74.03                                                           & 73.93                                                     & 44.84                                                        & 65.14                                                            & 75.66                          & 68.67                                                                              & 41.18                                                           & 9.30                                                             \\
Llama-3.1-8B-Instruct           & 8                              & 51.76                                              & 70.31          & 62.58          & 70.18                        & 53.55          & 50.49          & 54.49                            & 48.82                                   & 57.21                                                      & 61.85                                                           & 63.91                                                     & 39.17                                                        & 55.24                                                            & 56.72                          & 53.15                                                                              & 21.57                                                           & 8.86                                                             \\
glm-4-9b-chat                   & 9                              & 53.83                                              & 68.78          & 63.25          & 57.43                        & 59.84          & 47.96          & 43.69                            & 55.09                                   & 45.23                                                      & 57.51                                                           & 72.70                                                     & 51.63                                                        & 62.45                                                            & 68.79                          & 64.36                                                                              & 34.64                                                           & 7.86                                                             \\
Qwen2.5-14B-Instruct            & 14                             & 70.54                                              & 85.20          & 84.40          & 81.39                        & 85.98          & 68.94          & 76.46                            & 68.65                                   & 70.04                                                      & 80.19                                                           & 72.24                                                     & 60.24                                                        & 69.68                                                            & 78.06                          & 75.05                                                                              & 56.54                                                           & 15.57                                                            \\
DeepSeek-V2-Lite-Chat           & 16                             & 39.09                                              & 55.97          & 39.18          & 30.73                        & 27.08          & 33.01          & 45.02                            & 35.97                                   & 39.29                                                      & 62.19                                                           & 63.09                                                     & 33.02                                                        & 55.54                                                            & 49.83                          & 41.61                                                                              & 9.48                                                            & 4.42                                                             \\
DeepSeek-Coder-V2-Lite-Instruct & 16                             & 49.40                                              & 66.51          & 50.46          & 64.25                        & 51.74          & 55.02          & 61.75                            & 50.03                                   & 49.15                                                      & 60.18                                                           & 58.85                                                     & 37.51                                                        & 42.53                                                            & 51.17                          & 50.62                                                                              & 27.45                                                           & 13.20                                                            \\
QwQ-32B-Preview                 & 32                             & 77.17                                              & 88.07          & 91.22          & 89.75                        & \textbf{92.35} & 79.94          & 81.76                            & 81.37                                   & 67.32                                                      & \textbf{87.95}                                                  & 75.34                                                     & \textbf{73.12}                                               & 64.64                                                            & 66.15                          & 82.15                                                                              & 77.12                                                           & 36.47                                                            \\
Qwen2.5-32B-Instruct            & 32                             & 75.18                                              & 86.41          & 86.84          & 80.90                        & 90.49          & 79.43          & 79.82                            & 76.09                                   & 69.84                                                      & 79.73                                                           & 77.70                                                     & 66.44                                                        & \textbf{76.73}                                                   & 80.42                          & 82.25                                                                              & 67.32                                                           & 22.43                                                            \\
Qwen2.5-Coder-32B-Instruct      & 32                             & 71.34                                              & 78.72          & 80.46          & 88.62                        & 86.20          & 69.90          & 71.01                            & 73.03                                   & 60.78                                                      & 79.25                                                           & 76.19                                                     & 67.17                                                        & 69.29                                                            & 82.03                          & 74.89                                                                              & 62.75                                                           & 21.06                                                            \\
Llama-3.3-70B-Instruct          & 70                             & 73.67                                              & 89.39          & 84.86          & 83.39                        & 80.08          & 75.90          & 73.65                            & 73.38                                   & \textbf{73.97}                                             & 81.91                                                           & 84.02                                                     & 64.33                                                        & 73.36                                                            & 86.90                          & 80.62                                                                              & 60.13                                                           & 12.87                                                            \\
Qwen2.5-72B-Instruct            & 72                             & 74.80                                              & 80.31          & 86.35          & 82.70                        & 88.63          & 67.38          & 80.97                            & 74.85                                   & 68.95                                                      & 85.21                                                           & 85.83                                                     & 72.22                                                        & 73.10                                                            & 79.76                          & 81.90                                                                              & 60.78                                                           & 27.92                                                            \\
DeepSeek-V2.5-1210              & 236                            & 73.60                                              & 91.32          & 81.08          & 89.81                        & 77.91          & 73.71          & 82.83                            & 71.49                                   & 63.67                                                      & 80.86                                                           & \textbf{86.81}                                            & 62.88                                                        & 66.00                                                            & 83.94                          & 78.72                                                                              & 63.73                                                           & 22.84                                                            \\
DeepSeek-V3                     & 671                            & 77.03                                              & \textbf{92.84} & 88.72          & 90.70                        & 82.49          & 78.33          & 84.73                            & 81.72                                   & 65.00                                                      & 86.63                                                           & 85.75                                                     & 67.17                                                        & 69.91                                                            & 78.24                          & \textbf{85.83}                                                                     & 67.97                                                           & 26.44                                                            \\
DeepSeek-R1                     & 671                            & \textbf{80.14}                                     & 87.26          & \textbf{93.85} & \textbf{96.37}               & 90.49          & \textbf{86.43} & \textbf{87.59}                   & \textbf{87.91}                          & 65.69                                                      & 76.44                                                           & 77.50                                                     & 59.77                                                        & 65.41                                                            & \textbf{87.39}                 & 78.29                                                                              & \textbf{86.60}                                                  & \textbf{55.30}                                                   \\ \midrule
qwen-max-2024-09-19             & N/A                            & 73.49                                              & 84.72          & 84.76          & 77.60                        & 84.94          & 75.41          & 80.50                            & 72.64                                   & 69.96                                                      & 82.17                                                           & 74.66                                                     & 65.88                                                        & 78.36                                                            & 80.58                          & 80.18                                                                              & 63.40                                                           & 20.12                                                            \\
Claude-3-5-sonnet-20241022      & N/A                            & 81.69                                              & 89.00          & \textbf{90.23} & \textbf{92.59}               & \textbf{95.57} & \textbf{89.00} & \textbf{87.46}                   & \textbf{85.76}                          & 69.55                                                      & 87.56                                                           & 75.94                                                     & \textbf{68.97}                                               & 74.81                                                            & \textbf{92.05}                 & \textbf{87.94}                                                                     & 78.10                                                           & 42.43                                                            \\
gpt-4o-mini-2024-07-18          & N/A                            & 68.69                                              & 84.82          & 79.99          & 76.62                        & 81.13          & 78.83          & 69.85                            & 73.13                                   & 54.16                                                      & 81.47                                                           & 85.55                                                     & 56.71                                                        & 65.59                                                            & 70.50                          & 73.66                                                                              & 47.39                                                           & 19.70                                                            \\
gpt-4o-2024-11-20               & N/A                            & 78.21                                              & \textbf{89.59} & 86.26          & 87.42                        & 88.10          & 86.74          & 83.53                            & 83.05                                   & \textbf{72.20}                                             & \textbf{88.26}                                                  & 75.07                                                     & 67.32                                                        & 75.96                                                            & 85.31                          & 83.20                                                                              & 65.34                                                           & 34.06                                                            \\
o1-preview                      & N/A                            & \textbf{82.03}                                     & 87.98          & 88.52          & 90.57                        & 92.93          & 84.80          & 82.21                            & 82.88                                   & 71.45                                                      & 85.64                                                           & \textbf{85.79}                                            & 60.46                                                        & \textbf{78.40}                                                   & 76.93                          & 83.38                                                                              & \textbf{90.52}                                                  & \textbf{70.03}                                                   \\ \bottomrule
\end{tabular}
}
\caption{Performance of LLMs on 16 TableQA sub-tasks.}
\label{task-analysis-minor}
\end{table*}

\begin{table*}[t]
\small
\centering
\setlength{\abovecaptionskip}{0.12cm}    
\setlength{\belowcaptionskip}{-0.6cm}
\begin{tabular}{@{}lcccccccc@{}}
\toprule
Model                           & Size & Vertical       & Horizontal     & Matrix         & Concise        & Hierarchical   & Nested         & Others         \\ \midrule
Qwen2.5-7B-Instruct             & 7    & 66.26          & 66.10          & 64.97          & 57.12          & 61.65          & 53.04          & 61.07          \\
Llama-3.1-8B-Instruct           & 8    & 59.24          & 59.95          & 54.09          & 52.34          & 50.23          & 37.30          & 48.83          \\
glm-4-9b-chat                   & 9    & 57.95          & 58.86          & 60.76          & 53.53          & 53.03          & 56.36          & 57.64          \\
Qwen2.5-14B-Instruct            & 14   & 74.38          & 72.86          & 76.36          & 73.34          & 72.93          & 70.91          & 72.70          \\
DeepSeek-V2-Lite-Chat           & 16   & 43.05          & 22.81          & 39.48          & 38.49          & 40.08          & 34.46          & 44.10          \\
DeepSeek-Coder-V2-Lite-Instruct & 16   & 53.90          & 57.48          & 52.11          & 48.35          & 50.08          & 43.95          & 54.43          \\
QwQ-32B-Preview                 & 32   & 79.67          & 78.86          & 78.35          & 77.11          & 79.81          & 68.05          & 80.07          \\
Qwen2.5-32B-Instruct            & 32   & 80.99          & 75.05          & 79.60          & 75.81          & 76.59          & 71.53          & 71.26          \\
Qwen2.5-Coder-32B-Instruct      & 32   & 76.10          & 78.19          & 79.10          & 71.78          & 70.55          & 68.47          & 70.33          \\
Llama-3.3-70B-Instruct          & 70   & 80.54          & 77.43          & 78.15          & 75.53          & 73.11          & 65.83          & 72.94          \\
Qwen2.5-72B-Instruct            & 72   & 77.77          & \textbf{86.10} & \textbf{83.13} & 72.42          & 75.27          & 63.43          & 80.03          \\
DeepSeek-V2.5-1210              & 236  & 78.69          & 73.43          & 81.94          & 73.21          & 75.31          & \textbf{74.28} & 74.34          \\
DeepSeek-V3                     & 671  & 82.96          & 78.76          & 82.26          & 76.68          & 78.24          & 73.13          & 70.57          \\
DeepSeek-R1                     & 671  & \textbf{85.83} & 81.10          & 83.00          & \textbf{84.86} & \textbf{81.59} & 72.96          & \textbf{85.69} \\ \midrule
qwen-max-2024-09-19             & N/A  & 78.68          & 66.76          & 80.74          & 73.75          & 75.66          & 68.86          & 65.29          \\
Claude-3-5-sonnet-20241022      & N/A  & \textbf{86.61} & 74.43          & \textbf{87.59} & \textbf{84.77} & \textbf{83.03} & 71.05          & \textbf{84.27} \\
gpt-4o-mini-2024-07-18          & N/A  & 74.82          & 67.10          & 73.38          & 67.80          & 69.97          & 64.11          & 65.55          \\
gpt-4o-2024-11-20               & N/A  & 85.59          & \textbf{95.24} & 81.13          & 76.76          & 78.70          & 74.17          & 79.97          \\
o1-preview                      & N/A  & 84.96          & 77.00          & 83.81          & 84.40          & 82.71          & \textbf{75.46} & 83.60          \\ \bottomrule
\end{tabular}
\caption{Structure-aspect performance.}
\label{structure-level}
\end{table*}

Notably, while numerical analysis and data analysis tasks show slightly lower correlation coefficients (Pearson values of 0.8509 and 0.8972 respectively), these values remain substantially higher than those achieved by traditional metrics as shown in our main results. This slight decrease in correlation for quantitative tasks may reflect the inherent complexity of evaluating numerical reasoning, where small computational errors can lead to significant discrepancies in final answers. Nevertheless, the consistently high correlation values across all task types demonstrate SEAT's robustness as an evaluation framework regardless of the specific TableQA function being assessed.

\textbf{Domain-Specific Robustness.} \autoref{correlation-domain} extends our analysis to examine SEAT's performance across four domains from which tabular data originate. The results reveal SEAT's consistent effectiveness across various subject matters. Governmental data shows the highest correlation, likely due to the structured and standardized nature of such information. Both financial reports/company announcements and industry/stock research reports domains also demonstrate very high correlations (Pearson values of 0.9575 and 0.9605 respectively), indicating SEAT effectively evaluates responses in domains with specific terminology and numerical components.

Academic papers represent the most challenging domain, though the correlation values remain strong. This slight decrease may be attributed to the complexity and diversity of academic content, which often contains specialized vocabulary, complex methodology descriptions, and multi-faceted findings that can be challenging to evaluate in a structured format. Despite this relative difference, the consistently high correlation across all domains confirms that SEAT maintains its effectiveness regardless of the subject matter or data source.

These comprehensive experiments across both task types and domains provide strong evidence for SEAT's robustness as an evaluation framework.

\begin{table*}[t]
\small
\centering
\setlength{\abovecaptionskip}{0.12cm}    
\setlength{\belowcaptionskip}{-0.5cm}
\resizebox{\linewidth}{!}{
\begin{tabular}{@{}lccccccc@{}}
\toprule
\textbf{Model}        & \textbf{Avg} & \textbf{\begin{tabular}[c]{@{}c@{}}Information \\ Retrival\end{tabular}} & \textbf{\begin{tabular}[c]{@{}c@{}}Numerical \\ Analysis\end{tabular}} & \textbf{Reasoning} & \textbf{\begin{tabular}[c]{@{}c@{}}Data \\ Analysis\end{tabular}} & \textbf{\begin{tabular}[c]{@{}c@{}}Multi-turn \\ Conversation\end{tabular}} & \textbf{\begin{tabular}[c]{@{}c@{}}Table Structure\\ Understanding\end{tabular}} \\ \midrule
\multicolumn{8}{c}{\textbf{HTML}}                                                                                                                                                                                                                                                                                                                                                                                                                  \\ \midrule
gpt-4o-2024-11-20     & 81.09        & 89.98                                                                    & 87.26                                                                  & 80.92              & 79.80                                                             & 75.65                                                                       & 72.95                                                                            \\
Qwen2.5-14B-Instruct  & 72.51        & 80.93                                                                    & 79.85                                                                  & 73.21              & 73.17                                                             & 75.37                                                                       & 52.56                                                                            \\
Llama-3.1-8B-Instruct & 53.19        & 70.28                                                                    & 53.32                                                                  & 50.52              & 56.86                                                             & 56.87                                                                       & 31.32                                                                            \\ \midrule
\multicolumn{8}{c}{\textbf{Latex}}                                                                                                                                                                                                                                                                                                                                                                                                                 \\ \midrule
gpt-4o-2024-11-20     & 81.04        & 88.00                                                                    & 86.19                                                                  & 83.04              & 80.66                                                             & 82.95                                                                       & 65.38                                                                            \\
Qwen2.5-14B-Instruct  & 72.67        & 82.66                                                                    & 81.10                                                                  & 72.56              & 73.19                                                             & 76.83                                                                       & 49.69                                                                            \\
Llama-3.1-8B-Instruct & 54.16        & 69.51                                                                    & 56.90                                                                  & 52.77              & 56.85                                                             & 59.10                                                                       & 29.84                                                                            \\ \midrule
\multicolumn{8}{c}{\textbf{Markdown}}                                                                                                                                                                                                                                                                                                                                                                                                              \\ \midrule
gpt-4o-2024-11-20     & 78.79        & 88.24                                                                    & 86.00                                                                  & 83.05              & 81.47                                                             & 83.20                                                                       & 50.79                                                                            \\
Qwen2.5-14B-Instruct  & 70.02        & 84.72                                                                    & 78.93                                                                  & 68.65              & 75.06                                                             & 75.05                                                                       & 37.72                                                                            \\
Llama-3.1-8B-Instruct & 49.26        & 67.40                                                                    & 53.35                                                                  & 48.82              & 57.06                                                             & 53.15                                                                       & 15.76                                                                            \\ \bottomrule
\end{tabular}
}
\caption{Performance of LLMs on TableEval with different table formats.}
\label{format-main}
\end{table*}

\subsection{Performance on Sub-tasks}
\autoref{task-analysis-minor} provides a more granular evaluation of each model’s performance across 16 sub-tasks, offering more profound insight into LLMs' specific strengths and weaknesses. These sub-tasks include broader categories of \emph{information retrieval}, \emph{numerical analysis}, \emph{reasoning}, \emph{data analysis}, \emph{multi-turn conversation}, and \emph{table structure understanding}, each targeting a unique aspect of TableQA.

It can be found in the \emph{information retrieval} sub-tasks (simple, condition, and grouped lookups) that larger or more specialized models (particularly DeepSeek-R1, Claude 3.5 sonnet, GPT-4o, and QwQ-32B-Preview) consistently achieve the highest performance. QwQ-32B-Preview and DeepSeek-R1 better handle more complex filtering (condition, grouped), suggesting that the enhanced reasoning capabilities help manage varied data partitions. A similar pattern is shown in \emph{numerical analysis} (sorting, statistics, calculation), with Claude 3.5 sonnet exceeding 95\% on sorting and DeepSeek-R1 exceeding 86\% on statistics and calculation. This performance indicates that large model sizes and specialized data or method-specific fine-tuning can improve arithmetic-driven tasks.

For \emph{reasoning}, we focus on multi-hop questions requiring chained inference across table cells. DeepSeek-R1 achieves a top score of 87.91\%, followed by closed-source models such as Claude 3.5 sonnet and GPT-4o, highlighting the benefits of size and specialized training. Smaller open-source models (e.g., glm-4-9b-chat, DeepSeek-V2-Lite-Chat) fall behind, reflecting the complexity of multi-hop inference towards small-scale models.

\emph{Data analysis} includes six sub-tasks (causal analysis, comparative analysis, trend analysis, anomaly detection, correlation judgment, and rejection). These tasks require higher-level inference and domain knowledge, and the results reveal significant performance spread across models. For example, Llama-3.3-70B-Instruct achieves high scores in causal and trend analysis, while QwQ-32B-Preview achieves one of the highest scores on anomaly detection. Conversely, o1-preview demonstrates strong correlation judgment (78.40\%), and Claude 3.5 sonnet outperforms all other models on rejection (92.05\%), reflecting its robustness when queries demand domain knowledge or interpretation of numeric relationships. The difference in these data analysis tasks suggests that even leading LLMs specialize differently, implying opportunities for method-level customizations to boost performance on specific analytical tasks.

In the \emph{ellipsis/reference} sub-task, Claude 3.5 sonnet achieves the best (87.94\%), followed closely by DeepSeek-V3. Large-scale open-source models (e.g., Llama-3.3-70B-Instruct, Qwen2.5-72B-Instruct) also obtain high scores, yet they still fall behind top closed-source systems, demonstrating the benefit of extensive multi-turn dialogue training data in proprietary setups.

\begin{table*}[t]
\small
\centering
\setlength{\abovecaptionskip}{0.12cm}    
\setlength{\belowcaptionskip}{-0.3cm}
\resizebox{\linewidth}{!}{
\begin{tabular}{@{}lccccccccccccccccc@{}}
\toprule
\multirow{2}{*}{\textbf{Model}} & \multicolumn{1}{c|}{\multirow{2}{*}{\textbf{AVG}}} & \multicolumn{3}{c|}{\textbf{Information Retrival}} & \multicolumn{3}{c|}{\textbf{Numerical Analysis}}        & \multicolumn{1}{c|}{\textbf{Reasoning}} & \multicolumn{6}{c|}{\textbf{Data Analysis}}                                                                                                                                                                                                                                                                                                                 & \multicolumn{1}{c|}{\textbf{Conversation}}                                         & \multicolumn{2}{c}{\textbf{Table Structure Understanding}}                                                                         \\ \cmidrule(l){3-18} 
                                & \multicolumn{1}{c|}{}                              & Simple  & Condition & \multicolumn{1}{c|}{Grouped} & Sorting & Statistics & \multicolumn{1}{c|}{Calculation} & \multicolumn{1}{c|}{Multi-hop}          & \begin{tabular}[c]{@{}c@{}}Causal \\ Analysis\end{tabular} & \begin{tabular}[c]{@{}c@{}}Comparative \\ Analysis\end{tabular} & \begin{tabular}[c]{@{}c@{}}Trend \\ Analysis\end{tabular} & \begin{tabular}[c]{@{}c@{}}Anomaly \\ Detection\end{tabular} & \begin{tabular}[c]{@{}c@{}}Correlation \\ Judgement\end{tabular} & \multicolumn{1}{c|}{Rejection} & \multicolumn{1}{c|}{\begin{tabular}[c]{@{}c@{}}Ellipsis/\\ Reference\end{tabular}} & \begin{tabular}[c]{@{}c@{}}Table Size \\ Detection\end{tabular} & \begin{tabular}[c]{@{}c@{}}Merged Cell \\ Detection\end{tabular} \\ \midrule
\multicolumn{18}{c}{\textbf{HTML}}                                                                                                                                                                                                                                                                                                                                                                                                                                                                                                                                                                                                                                                                                                                                                                                                    \\ \midrule
gpt-4o-2024-11-20               & 80.29                                              & 90.10   & 89.20     & 94.94                        & 91.86   & 89.32      & 81.63                            & 80.92                                   & 63.84                                                      & 81.52                                                           & 76.94                                                     & 63.36                                                        & 70.96                                                            & 88.23                          & 75.65                                                                              & 71.35                                                           & 74.78                                                            \\
Qwen2.5-14B-Instruct            & 71.87                                              & 79.60   & 82.73     & 83.07                        & 85.37   & 72.84      & 77.44                            & 73.21                                   & 62.05                                                      & 76.07                                                           & 74.94                                                     & 65.09                                                        & 60.97                                                            & 78.62                          & 75.37                                                                              & 69.61                                                           & 32.94                                                            \\
Llama-3.1-8B-Instruct           & 54.98                                              & 71.30   & 68.66     & 70.75                        & 54.83   & 48.72      & 53.93                            & 50.52                                   & 63.16                                                      & 61.85                                                           & 71.44                                                     & 39.67                                                        & 53.05                                                            & 53.94                          & 56.87                                                                              & 41.57                                                           & 19.34                                                            \\ \midrule
\multicolumn{18}{c}{\textbf{Latex}}                                                                                                                                                                                                                                                                                                                                                                                                                                                                                                                                                                                                                                                                                                                                                                                                   \\ \midrule
gpt-4o-2024-11-20               & 79.51                                              & 87.66   & 88.54     & 87.82                        & 92.76   & 84.80      & 80.15                            & 83.04                                   & 70.04                                                      & 80.69                                                           & 79.03                                                     & 63.94                                                        & 72.51                                                            & 88.17                          & 82.95                                                                              & 70.05                                                           & 59.93                                                            \\
Qwen2.5-14B-Instruct            & 72.42                                              & 82.88   & 82.03     & 85.19                        & 85.17   & 76.77      & 78.93                            & 72.56                                   & 58.25                                                      & 79.61                                                           & 80.36                                                     & 64.61                                                        & 61.67                                                            & 76.86                          & 76.83                                                                              & 64.38                                                           & 32.67                                                            \\
Llama-3.1-8B-Instruct           & 54.98                                              & 71.18   & 67.32     & 66.46                        & 59.40   & 54.55      & 55.48                            & 52.77                                   & 55.88                                                      & 62.21                                                           & 69.56                                                     & 42.63                                                        & 49.84                                                            & 56.28                          & 59.10                                                                              & 46.73                                                           & 10.27                                                            \\ \midrule
\multicolumn{18}{c}{\textbf{Markdown}}                                                                                                                                                                                                                                                                                                                                                                                                                                                                                                                                                                                                                                                                                                                                                                                                \\ \midrule
gpt-4o-2024-11-20               & 78.21                                              & 89.59   & 86.26     & 87.42                        & 88.10   & 86.74      & 83.53                            & 83.05                                   & 72.20                                                      & 88.26                                                           & 75.07                                                     & 67.32                                                        & 75.96                                                            & 85.31                          & 83.20                                                                              & 65.34                                                           & 34.06                                                            \\
Qwen2.5-14B-Instruct            & 70.54                                              & 85.20   & 84.40     & 81.39                        & 85.98   & 68.94      & 76.46                            & 68.65                                   & 70.04                                                      & 80.19                                                           & 72.24                                                     & 60.24                                                        & 69.68                                                            & 78.06                          & 75.05                                                                              & 56.54                                                           & 15.57                                                            \\
Llama-3.1-8B-Instruct           & 51.76                                              & 70.31   & 62.58     & 70.18                        & 53.55   & 50.49      & 54.49                            & 48.82                                   & 57.21                                                      & 61.85                                                           & 63.91                                                     & 39.17                                                        & 55.24                                                            & 56.72                          & 53.15                                                                              & 21.57                                                           & 8.86                                                             \\ \bottomrule
\end{tabular}
}
\caption{Performance of LLMs on 16 sub-tasks with different table formats.}
\label{format-minor}
\end{table*}

\begin{table*}[t]
\setlength{\abovecaptionskip}{0.12cm}    
\setlength{\belowcaptionskip}{-0.6cm}
\resizebox{\linewidth}{!}{
\begin{tabular}{@{}lccccccc@{}}
\toprule
\textbf{Model}                     & \textbf{Avg} & \textbf{\begin{tabular}[c]{@{}c@{}}Information \\ Retrival\end{tabular}} & \textbf{\begin{tabular}[c]{@{}c@{}}Numerical \\ Analysis\end{tabular}} & \textbf{Reasoning} & \textbf{\begin{tabular}[c]{@{}c@{}}Data \\ Analysis\end{tabular}} & \textbf{\begin{tabular}[c]{@{}c@{}}Multi-turn \\ Conversation\end{tabular}} & \textbf{\begin{tabular}[c]{@{}c@{}}Table Structure \\ Understanding\end{tabular}} \\ \midrule
gpt-4o-mini-2024-07-18             & 68.47        & 82.64                                                                    & 76.15                                                                  & 73.13              & 70.70                                                             & 73.66                                                                       & 34.56                                                                             \\
Qwen2.5-7B-Instruct (fine-tuned)   & 66.19        & 80.06                                                                    & 61.71                                                                  & 58.73              & 70.56                                                             & 70.92                                                                       & 55.16                                                                             \\
glm-4-9b-chat (fine-tuned)         & 57.62        & 76.32                                                                    & 48.57                                                                  & 45.28              & 66.67                                                             & 62.60                                                                       & 46.30                                                                             \\
Llama-3.1-8B-Instruct (fine-tuned) & 63.83        & 79.88                                                                    & 60.09                                                                  & 53.31              & 68.84                                                             & 71.71                                                                       & 49.18                                                                             \\ \midrule
Qwen2.5-7B-Instruct                & 59.60        & 69.23                                                                    & 64.29                                                                  & 59.38              & 69.71                                                             & 68.67                                                                       & 26.35                                                                             \\
glm-4-9b-chat                      & 53.61        & 66.19                                                                    & 51.09                                                                  & 55.09              & 62.47                                                             & 64.36                                                                       & 22.44                                                                             \\
Llama-3.1-8B-Instruct              & 49.26        & 67.40                                                                    & 53.35                                                                  & 48.82              & 57.06                                                             & 53.15                                                                       & 15.76                                                                             \\ \bottomrule
\end{tabular}
}
\caption{Performance of fine-tuned LLMs on TableEval.}
\label{tab:finetune}
\end{table*}

Finally, \emph{table structure understanding} is assessed via table size detection and merged cell detection, which reflect the ability to parse and interpret structural details. While some models perform well on table size detection (led by o1-preview at 90.52\%), performance on merged cell detection drops significantly; no model exceeds 70.03\%, also achieved by o1-preview. This large gap highlights the challenge of complex structural tasks, where even advanced LLMs struggle to accurately capture and interpret non-trivial table structures (layouts) or spanning cell configurations.

Overall, the fine-grained results demonstrate the multi-dimensional capability of LLMs in tabular data. Although large models dominate most categories, tasks such as merged cell detection or advanced data analysis remain challenging. These observations highlight the importance of structure-aware designs, targeted fine-tuning for analytical tasks, and more elaborate training on real-world tables (particularly those with nested or merged cells). By bridging these gaps, LLMs will be better equipped to handle the broad tasks of TableQA.

\subsection{Impact of Table Format} \label{sec:format}

We conduct an experiment to explore the impact of table format on LLM performance by converting tables to LaTeX, HTML, and Markdown formats. We report the main results and the fine-grained results of 3 LLMs on \autoref{format-main} and \autoref{format-minor}.

We find that all three models achieve approximately 2 percentage points higher F1 scores on HTML and Latex formats than Markdown. We further look into \autoref{format-minor}: the main improvement comes from the table structure understanding task, where all models significantly improve F1 scores. Specifically, GPT-4o shows a 22.16\% and 14.59\% increase in F1 scores on HTML and LaTeX tables, respectively, compared to Markdown. The improvement is even more significant for Llama-3.1-8B-Instruct, with nearly double the performance gain.

In sub-task evaluations, all models exhibit around a 2$\times$ improvement in merged cell detection on HTML tables. Additionally, in table size detection, all models show varying degrees of progress. These results indicate that structured table formats (HTML and LaTeX) significantly enhance the models' ability to understand table structures, leading to improved TableQA performance.

\subsection{Impact of Fine-tuning}

We initially obtained 34,161 distinct questions, from which 2,325 were selected as the final dataset (\S\ref{construction}). Based on the remaining questions, we applied the QA acquisition process described in \S\ref{construction} to generate QA pairs. These pairs undergo a consistency check but do not undergo a human review. Finally, we select 16,844 QA pairs for LLM fine-tuning. The results are presented in \autoref{tab:finetune}.

Our experimental results demonstrate that fine-tuning significantly enhances the TableQA capabilities of LLMs. Even small-scale LLMs can achieve comparable performance with GPT-4o-mini after fine-tuning. Specifically, the fine-tuned Qwen2.5-7B-Instruct and Llama-3.1-8B-Instruct achieved notable improvements, particularly in table structure understanding, where they even outperform GPT-4o-mini. These results suggest that fine-tuning with task-specific data can help models develop a more structured understanding of tables.

Despite these advancements, challenges remain, particularly in numerical reasoning and complex analytical tasks, where even the fine-tuned models still fall behind. While information retrieval and data analysis have improved, the ability to perform deep numerical computations and multi-step reasoning over structured tables continues to be a key limitation. These findings indicate that further research is needed to enhance models' structured reasoning abilities, potentially through structure-aware training, hybrid retrieval-based approaches, or integration with external knowledge sources.

Overall, our study highlights the importance of fine-tuning for improving TableQA performance, demonstrating that with sufficient task-specific adaptation, small-scale open-source LLMs can achieve competitive results against stronger proprietary baselines. In the future work, we can focus on refining structured comprehension and enhancing numerical reasoning

\section{Prompts}

%\noindent\begin{minipage}[H]{0.48 \textwidth} % 宽度设置为整个文本宽度
%\begin{CJK}{UTF8}{gbsn}
%\begin{prompt}[title={Prompt (Template-prompted Strategy)}]
%\begin{verbatim}
%# 任务： 
%请你根据以下表格，将【简单查询、条件查询、分组查询、排序、统计、数值计算、因果分析、对比分析、趋势分析、异常检测、相关性判断、拒答、多轮问答】中的每个任务提出1个中文问题，并给出相应的答案。
%- 注意你提出的所有的问题都是客观问题，能从表格中得出答案。
%- 问题需贴近用户真实可能会问的问题。
%- 提出的问题需要提及到表格的名称等信息，避免产生表格的歧义。
%
%## 任务描述：
%- 简单查询：针对单一字段或多个字段的查询，不涉及任何条件过滤、数据比较或者计算。包括直接从表格中获取的事实性信息和解释。
%- 条件查询：根据特定字段（如时间、数值范围、类别）对表格数据进行筛选，获取满足条件的数据。
%- 分组查询：根据特定字段对数据进行分组聚合。
%- 排序：根据特定指标对表格中的数据进行排序或求极值，包括查最大值、最小值、前几名或后几名的结果。
%- 统计：对表格中的数据进行简单的计数，不涉及复杂的数学运算或多步推理。
%- 数值计算：对表格中的数据进行数学运算，得到具体的数值结果，关注计算过程和结果。这包括计算平均值、中位数、和、差值、乘积、比例、范围等。
%- 因果分析：对数据之间的关系进行分析，确定哪些因素对目标变量产生影响，或解释某个结果或变化背后的原因。通常需要结合多个数据点，可能涉及定量评估和定性解释。
%- 对比分析：比较两个或多个数据点、对象、时间点之间的差异或相似性，主要是定性分析。
%- 趋势分析：涉及连续分析数据随时间的变化，关注长期趋势和变化模式。
%- 异常检测：识别数据中的异常值或不寻常的模式。
%- 相关性判断：判断根据表格内容，是否可以回答该问题。
%- 拒答：仅根据表格内容无法回答的问题。
%- 多轮问答：针对表格内容进行多轮对话，你需要提出多个问题，下一个问题与上一个相关，问题涉及省略或指代。
%
%## 输出格式：以下方的JSON格式呈现，多轮问答的问题和答案采用列表形式：
%[
%    {
%      "问题": "" ,
%      "答案": ""
%      "类别": ""
%    },
%    {
%      "问题": "",
%      "答案": "",
%      "类别": ""
%    },
%    ...,
%    {
%      "问题": ["问题1", "问题2"],
%      "答案": ["答案1", "答案2"],
%      "类别": "多轮问答"
%    }
%]
%
%# 表格：
%{table}
%\end{verbatim}
%\end{prompt}
%\end{CJK}
%\vspace{-12pt} % 在标题和下文之间添加额外的间隔
%\captionof{figure}{QA generation prompt in template-prompted strategy.}
%\label{strategy1.1}
%%\vspace{10pt} % 在标题和下文之间添加额外的间隔
%\end{minipage}

\begin{figure}[h]
  \centering 
  \setlength{\abovecaptionskip}{0.12cm}    
  \setlength{\belowcaptionskip}{-0.6cm}
  \includegraphics[width=0.74\linewidth]{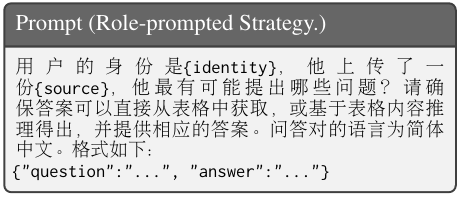}\\
  \caption{QA generation prompt in role-prompted strategy. \label{strategy2}}
\end{figure}

\begin{figure}[!htb]
  \centering 
  \setlength{\abovecaptionskip}{0.12cm}    
  \setlength{\belowcaptionskip}{-0.6cm}
  \includegraphics[width=0.83\linewidth]{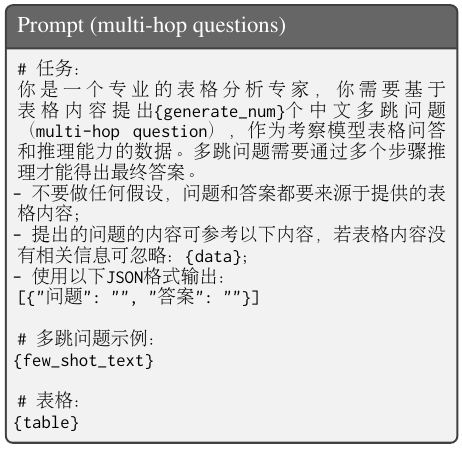}\\
  \caption{QA generation prompt for multi-hop questions in template-prompted strategy. \label{strategy1.2}}
\end{figure}

\begin{figure}[!htb]
  \centering 
  \setlength{\abovecaptionskip}{0.12cm}    
  \setlength{\belowcaptionskip}{-0.6cm}
  \includegraphics[width=0.83\linewidth]{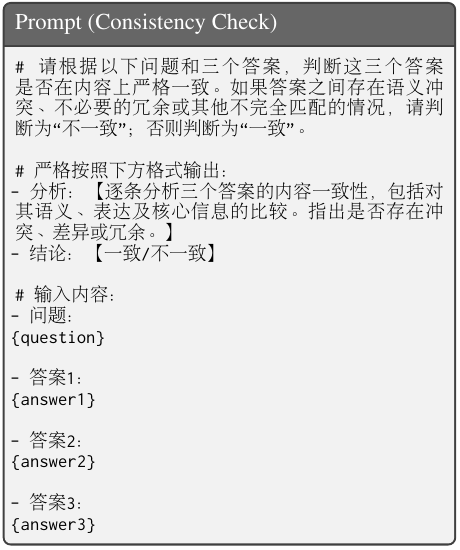}\\
  \caption{Consistency Check prompt. \label{consistency}}
\end{figure}

\begin{figure}[!htb]
  \centering 
  \setlength{\abovecaptionskip}{0.12cm}    
  \setlength{\belowcaptionskip}{-0.6cm}
  \includegraphics[width=0.83\linewidth]{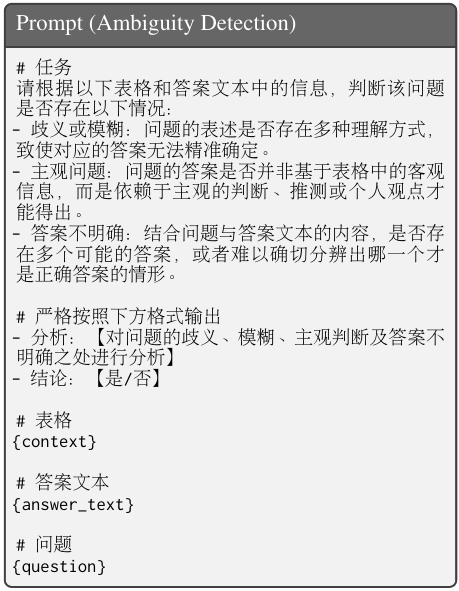}\\
  \caption{Ambiguity Detection prompt. \label{ambiguity}}
\end{figure}

\begin{figure}[h]
  \centering 
  \setlength{\abovecaptionskip}{0.12cm}    
  \setlength{\belowcaptionskip}{-0.4cm}
  \includegraphics[width=0.95\linewidth]{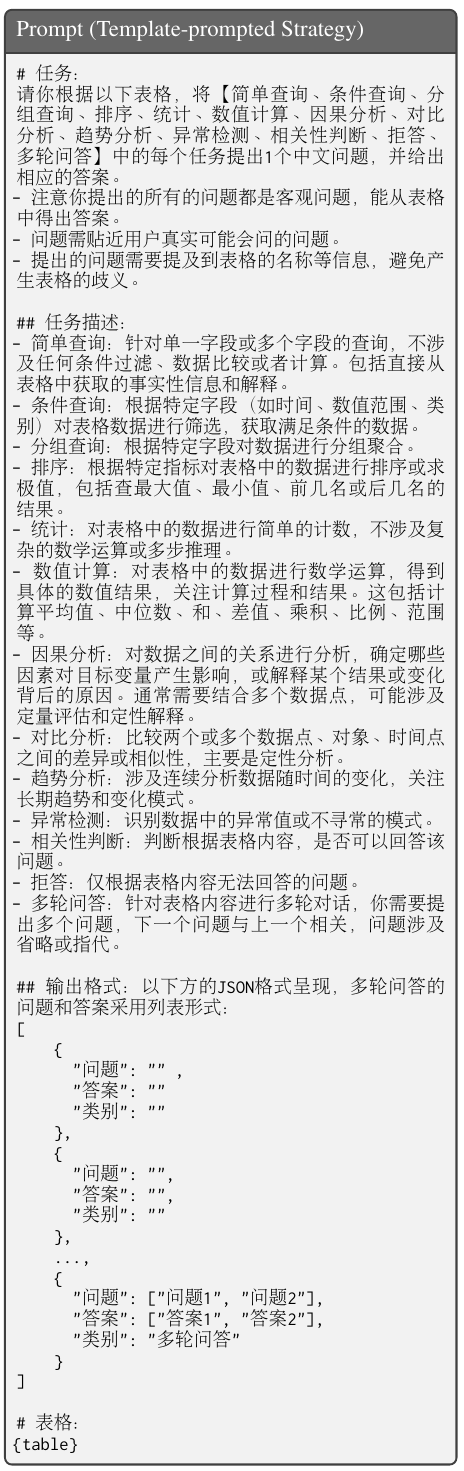}\\
  \caption{QA generation prompt in template-prompted strategy. \label{strategy1.1}}
\end{figure}

%\noindent\begin{minipage}[H]{0.48 \textwidth} % 宽度设置为整个文本宽度
%\begin{CJK}{UTF8}{gbsn}
%\begin{prompt}[title={Prompt (multi-hop questions)}]
%\begin{verbatim}
%# 任务：
%你是一个专业的表格分析专家，你需要基于表格内容提出{generate_num}个中文多跳问题（multi-hop question），作为考察模型表格问答和推理能力的数据。多跳问题需要通过多个步骤推理才能得出最终答案。
%- 不要做任何假设，问题和答案都要来源于提供的表格内容；
%- 提出的问题的内容可参考以下内容，若表格内容没有相关信息可忽略：{data}；
%- 使用以下JSON格式输出：
%[{"问题": "", "答案": ""}]
%
%# 多跳问题示例：
%{few_shot_text}
%
%# 表格：
%{table}
%\end{verbatim}
%\end{prompt}
%\end{CJK}
%\vspace{-12pt} % 在标题和下文之间添加额外的间隔
%\captionof{figure}{QA generation prompt for multi-hop questions in template-prompted strategy.}
%\label{strategy1.2}
%%\vspace{10pt} % 在标题和下文之间添加额外的间隔
%\end{minipage}

\begin{figure}[h]
  \centering 
  \setlength{\abovecaptionskip}{0.12cm}    
  \setlength{\belowcaptionskip}{-0.4cm}
  \includegraphics[width=0.95\linewidth]{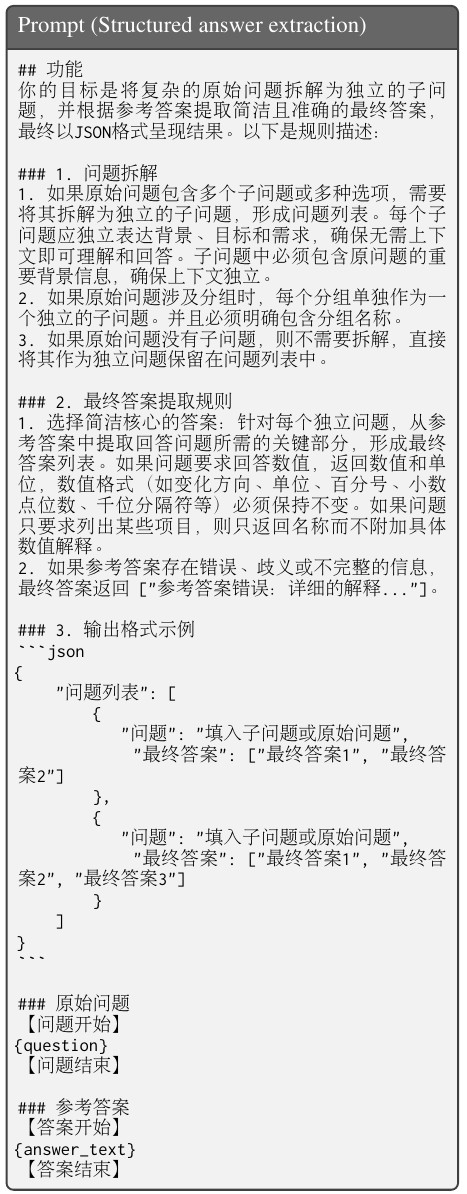}\\
  \caption{Structured answer extraction prompt. \label{structured}}
\end{figure}

\begin{figure}[h]
  \centering 
  \setlength{\abovecaptionskip}{0.12cm}    
  \setlength{\belowcaptionskip}{-0.4cm}
  \includegraphics[width=0.9\linewidth]{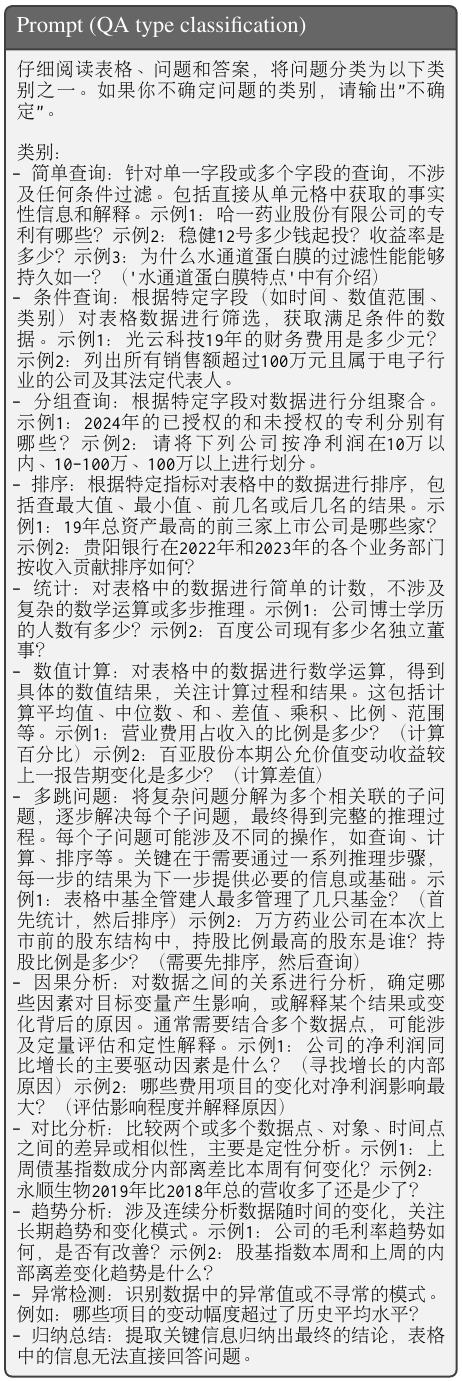}\\
  \caption{QA type classification prompt used in role-prompted strategy. \label{qa_classification}}
\end{figure}

\begin{figure}[h]
  \centering 
  \setlength{\abovecaptionskip}{0.12cm}    
  \setlength{\belowcaptionskip}{-0.4cm}
  \includegraphics[width=0.95\linewidth]{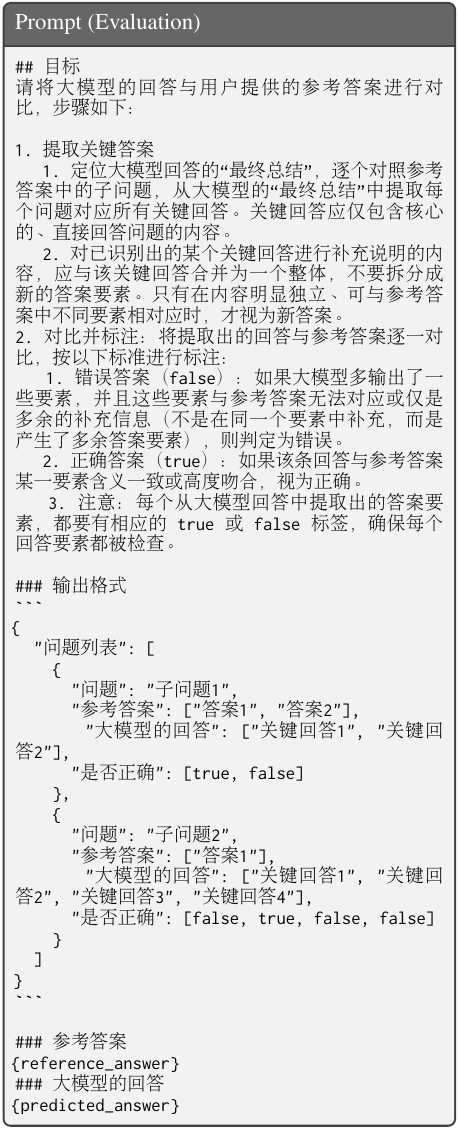}\\
  \caption{Evaluation prompt in SEAT. \label{evaluation_prompt}}
\end{figure}

\begin{figure}[t]
  \centering 
  \setlength{\abovecaptionskip}{0.12cm}    
  \setlength{\belowcaptionskip}{-0.4cm}
  \includegraphics[width=0.95\linewidth]{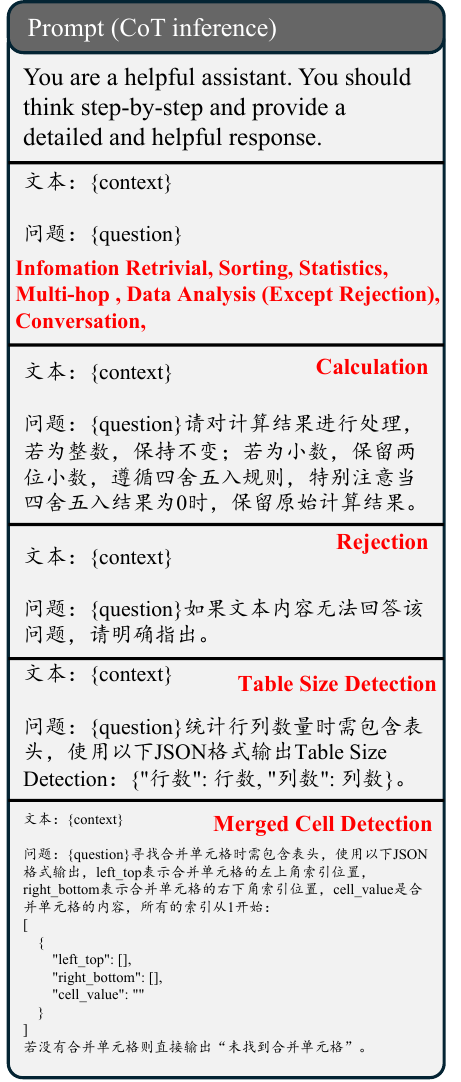}\\
  \caption{CoT prompt in inference. The prompt varies depending on the tasks, with the text in red representing the tasks corresponding to each \label{cot_prompt}}
\end{figure}

\begin{figure*}[t]
  \centering 
  \includegraphics[width=5.5 in]{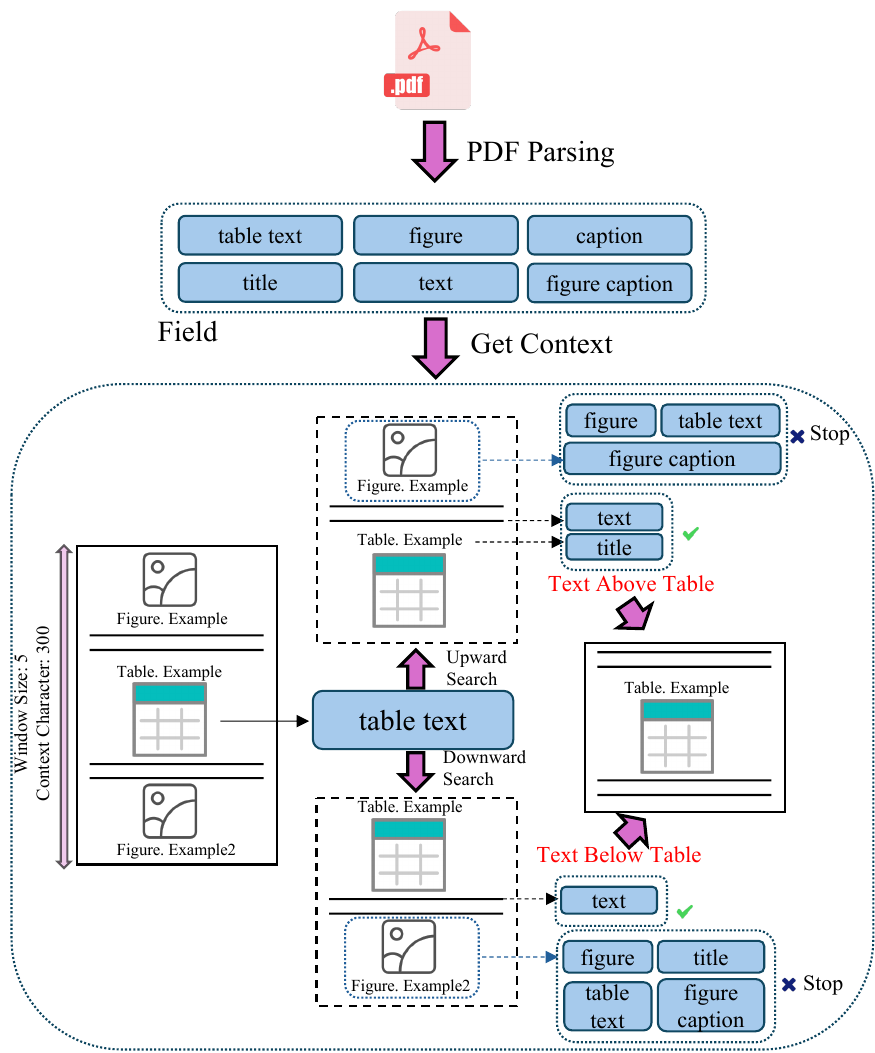}\\
  \caption{Overview of Heuristic Rules for Table Context Extraction. \label{rule}}
\end{figure*}

\begin{algorithm*}[t]
\small
\caption{Heuristic Rules for Table Context Extraction}
\label{alg:table_context_extraction}
\textbf{Input:} 
\begin{itemize}
    \item $\mathcal{L}$: PDF layout analysis results, a list of elements, including layout type and textual content.
    \item $\omega$: Window range for context extraction, expressed as the number of paragraphs or layout blocks.
    \item $\kappa$: Character count limit for the extracted context.
\end{itemize}

\textbf{Output:} 
\begin{itemize}
    \item $\mathcal{T}$: A set of tables, each table augmented with its extracted upward and downward context.
\end{itemize}

\begin{algorithmic}[1]
\Procedure{ExtractTableContext}{$\mathcal{L}, \omega, \kappa$}
    \State $\mathcal{T} \gets \texttt{LocateTables}(\mathcal{L})$ \Comment{Identify all tables in the layout}
    \For{each Table $T \in \mathcal{T}$}
        \State $C_{\text{up}} \gets \emptyset$ \Comment{Initialize upward context}
        \State $C_{\text{down}} \gets \emptyset$ \Comment{Initialize downward context}

        \Comment{Upward Context Search}
        \For{each Window $w$ in \texttt{TraverseUpward}($T, \omega$)}
            \If{\texttt{IsHeader}($w$)}
                \State $C_{\text{up}} \gets \texttt{ExtractText}(w, T)$
                \State \textbf{break}
            \ElsIf{\texttt{IsNonTextContent}($w$)}
                \State $C_{\text{up}} \gets \texttt{ExtractTextBelow}(w)$
                \State \textbf{break}
            \ElsIf{\texttt{CharCountExceeded}($C_{\text{up}}, \kappa$)}
                \State \textbf{break}
            \ElsIf{\texttt{WindowRangeExceeded}($w$)}
                \State \textbf{break}
            \EndIf
            \State \texttt{SkipIrrelevantContent}($w$)
        \EndFor

        \Comment{Downward Context Search}
        \For{each Window $w$ in \texttt{TraverseDownward}($T, \omega$)}
            \If{\texttt{IsHeader}($w$)}
                \State $C_{\text{down}} \gets \texttt{ExtractText}(T, w)$
                \State \textbf{break}
            \ElsIf{\texttt{IsNonTextContent}($w$)}
                \State $C_{\text{down}} \gets \texttt{ExtractTextAbove}(w)$
                \State \textbf{break}
            \ElsIf{\texttt{CharCountExceeded}($C_{\text{down}}, \kappa$)}
                \State \textbf{break}
            \ElsIf{\texttt{WindowRangeExceeded}($w$)}
                \State \textbf{break}
            \EndIf
            \State \texttt{SkipIrrelevantContent}($w$)
        \EndFor

        \State \texttt{CombineContext}($T, C_{\text{up}}, C_{\text{down}}$)
    \EndFor

    \State \Return $\mathcal{T}$ \Comment{Return tables with augmented context}
\EndProcedure

\Function{LocateTables}{$\mathcal{L}$}
    \State \Return \texttt{ElementsOfType}($\mathcal{L}, \text{``Table''}$)
\EndFunction

\end{algorithmic}
\end{algorithm*}

\end{document}